# A survey on machine reading comprehension systems

Razieh Baradaran[1,*], Razieh Ghiasi[1,*], and Hossein Amirkhani[1,**]

[1] Computer and Information Technology Department, University of Qom, Qom, Iran
[*] Equal contribution
[**] Corresponding author. Email: amirkhani@qom.ac.ir

## Abstract

Machine reading comprehension is a challenging task and hot topic in natural language processing. Its goal is to develop systems to answer the questions regarding a given context. In this paper, we present a comprehensive survey on different aspects of machine reading comprehension systems, including their approaches, structures, input/outputs, and research novelties. We illustrate the recent trends in this field based on 241 reviewed papers from 2016 to 2020. Our investigations demonstrate that the focus of research has changed in recent years from answer extraction to answer generation, from single to multi-document reading comprehension, and from learning from scratch to using pre-trained embeddings. We also discuss the popular datasets and the evaluation metrics in this field. The paper ends with investigating the most cited papers and their contributions.

## 1. Introduction

The task of machine reading comprehension (MRC) is a useful benchmark to evaluate natural language understanding of machines and has been a challenging task in natural language processing (NLP) field with considerable research in recent years. For measuring the machine comprehension of a piece of natural language text, a set of questions about the text is given to the machine, and its responses are evaluated against the gold standard. According to Burgess (2013), a machine is considered to understand a text if, for any question about the text that can be answered by a majority of native speakers, it provides an answer with two characteristics: it is in comply with the humans' answers and it does not include irrelevant information (Burges 2013). Also, based on Chen et al., machine reading comprehension is the machine's ability to read a text, process it, understand its meaning, and answer related questions (Chen, Bolton, and Manning 2016). Each instance in MRC datasets contains a context $C$, a related question $Q$, and the answer $A$. Figure 1 shows some examples of the SQuAD (Rajpurkar et al. 2016) and CNN/Daily Mail (Hermann et al. 2015) datasets. The goal of MRC systems is to learn the predictive function $f$, which extracts/generates the appropriate answer $A$ by receiving the context $C$ and the related question $Q$:

$$f: (C, Q) \rightarrow A$$

Also, MRC systems have important applications in different areas such as conversational agents (Reddy, Chen, and Manning 2019;

Hewlett, Jones, and Lacoste 2017) and customer service support (Cui, Huang, et al. 2017).

| SQuAD dataset | CNN/Daily Mail dataset |
|---|---|
| **Context**<br>Alyssa got to the beach after a long trip. She's from Charlotte. She traveled from Atlanta. She's now in Miami. She went to Miami to visit some friends. But she wanted some time to herself at the beach, so she went there first. After going swimming and laying out, she went to her friend Ellen's house. Ellen greeted Alyssa and they both had some lemonade to drink. Alyssa called | **Context**<br>(@entity4) if you feel a ripple in the force today, it may be the news that the official @entity6 is getting its first gay character. according to the sci-fi website @entity9, the upcoming novel" @entity11" will feature a capable but flawed @entity13 official named @entity14 who" also happens to be a lesbian." The character is the first gay figure in the official @entity6 – the movies, television shows, comics and books approved by @entity6 franchise owner @entity22 -- according to @entity24, editor of "@entity6" books at @entity28 imprint @entity26. |
| **Question**<br>Why did Alyssa go to Miami? | **Question**<br>characters in "@placeholder" movies have gradually become more diverse. |
| **Answer**<br>To visit some friends | **Answer**<br>@entity6 |

**Figure 1**. Samples from SQuAD and CNN/Daily Mail datasets

Even though in some studies, MRC is referred to as question answering (QA), they are different in the following ways:

- The main objective of QA systems is to answer the input questions, while in an MRC system, as its name indicates, the main goal is to understand natural languages by machines.
- The only input to QA systems is the question, while the inputs to MRC systems are the question and the corresponding context that should be used to answer the question. For this reason, sometimes MRC is referred to as QA from text (Deng and Liu 2018).
- The main information source that is used to answer questions in MRC systems are natural language texts; while in QA systems, the structured and semi-structured data sources such as knowledge-bases are commonly used besides the non-structured data like texts.

The history of MRC dates back to the 1970s when researchers identified it as a convenient way to test computer comprehension ability (Chen 2018). One of the most prominent early studies was the QUALM system (Lehnert 1977). This system was limited to handwritten scripts and could not be easily generalized to larger domains. Due to the complexity of this task, research in this area was reduced in 1980s and 1990s. In the late 1990s, Hirschman et al. (1999) revived the field of MRC by creating a new dataset, including 120 stories and questions from 3rd to 6th-grade material, followed by a workshop on comprehension test as a tool for assessing machine comprehension at ANLP/NAACL 2000.

Another revolution in this field occurred between 2013 and 2015 by introducing labeled traning datasets mapping (context,question) pairs to the answer. This transformed the MRC problem into a supervised learning task (Chen 2018). Two prominent datasets in this period were the MCTest dataset (Richardson, Burges, and Renshaw 2013) with 500 stories and 2000 questions and the ProcessBank dataset (Berant et al. 2014) with 585 questions over 200 paragraphs related to biological processes. In 2015, the introduction of large datasets such as CNN / Daily Mail (Hermann et al. 2015) and SQuAD (Rajpurkar et al. 2016) opened a new window in the MRC field by allowing the development of deep models.

In recent years, with the success of machine learning techniques, especially the neural networks, and the usage of recurrent neural networks to process sequential data such as texts, MRC has become an active area in the field of NLP. The goal of this paper is to categorize these studies, provide related statistics, and show the trends in this field. Some recent surveys focused on QA systems (Bouziane et al. 2015; Kodra and Meçe 2017). Some papers presented a partial survey on some MRC systems but did not provide a comprehensive classification of different aspects and different statistics in this field (Arivuchelvan and Lakahmi 2017; Zhang, Yang, et al. 2019). Ingale and Singh only provided a review on MRC datasets (Ingale and Singh 2019). Liu et al. (Liu, Zhang, Zhang, Wang, et al. 2019) provide a review on different aspects of neural MRC models including definitions, differences, popular datasets, architectures, and new trends based on 85 papers. In our study, we present a comprehensive review on 241 papers, analyzing and categorizing MRC studies from different aspects including problem-solving approaches, input/output, model structures, research novelties, and datasets. We also provide statistics on the amount of research attention to these aspects in different years, which are not provided in previous reviews.

In order to select papers, the queries “reading comprehension”, “machine reading”, “machine reading comprehension”, and “machine comprehension” were submitted to the Google Scholar service[1]. Also, the ACL Anthology website[2], which includes top related NLP conferences such as *ACL*, *EMNLP*, *NAACL*, and *CoNLL* were searched with the same queries to extract the remaining related papers. We excluded the retrieved papers that were published only on *arXiv*. We also excluded the papers with the *conversational* or *dialogue MRC* as subjects as well as the QA papers with no novelty in the MRC phase. We limited our study to the papers published in recent years, i.e., from 2016 to September 2020. Table 1 shows the number of reviewed papers over different years.

[1] https://scholar.google.com/
[2] https://www.aclweb.org/anthology/

The contributions of this paper are as follows:

- Investigating recently published MRC papers from different perspectives including problem-solving approaches, system input/outputs, contributions of these studies, and evaluation metrics.
- Providing statistics for each category over different years and highlighting the trends in this field.
- Reviewing available datasets and classifying them based on important factors.
- Investigating the most cited papers from different aspects.

The rest of this paper is organized as follows. Section 2 focuses on the main problem-solving approaches for the MRC task. Section 3 provides an analysis of the type of input/outputs of MRC systems. The review of the papers based on the basic phases of an MRC system is presented in Section 4. The recent datasets and evaluation measures are reviewed in Sections 5 and 6, respectively. In Section 7, the MRC studies are categorized based on their contributions and novelties. The most cited papers are investigated in Section 8. Section 9 provides future trends and opportunities. Finally, the paper is concluded in Section 10.

**Table 1.** Number of reviewed papers over different years.

| YEAR | NUMBER OF PAPERS |
|---|---|
| 2016 | 13 |
| 2017 | 34 |
| 2018 | 62 |
| 2019 | 83 |
| 2020 | 49 |
| Total | 241 |

## 2. Problem-solving approaches

The approaches used for developing MRC systems can be grouped into three categories: rule-based methods, classical machine learning-based methods, and deep learning-based methods.

The traditional rule-based methods use the rules handcrafted by linguistic experts. These methods suffer from the problem of the incompleteness of the rules. Also, this approach is domain specific where for any new domain, a new set of rules should be handcrafted. As an example, Riloff and Thelen (2000) present a rule-based MRC system called Quarc, which reads a short story and answers the input question by extracting the most relevant sentences. Quarc uses a separate set of rules for each question type (WHO, WHAT, WHEN, WHERE, and WHY). In this system, several NLP tools are used for parsing, part of speech tagging, morphological analysis, entity recognition, and semantic class tagging. As another example, Akour et al. (2011) introduce the QArabPro system, which is a system for answering reading comprehension questions in the Arabic language. It is also developed using a set of rules for each type of question and uses multiple NLP components, including question classification, query reformulation, stemming, and root extraction.

The second approach is based on the classical machine learning. These methods rely on a set of human-defined features and train a model for mapping input features to the output. Note that in classical machine learning-based methods, even though the hand-crafted rules are not necessary, feature engineering is a critical necessity.

For example, Ng, Teo, and Kwan (2000) have developed a machine learning based MRC system and introduced some of features to be extracted from a context sentence like “the number of matching words/verb-types between the question and the sentence”, “the number of matching words/verb-types between the question and the previous/next sentence”, “co-reference information”, and binary features like “sentence-contain-person”, “sentence-contain-time”, “sentence-contain-location”, “sentence-is-title” and so on

The third approach uses deep learning methods to learn features from raw input data automatically. These methods require a large amount of training data to create high accuracy models. Because of the growth of available data and computational power in recent years, deep learning methods have gained state-of-the-art results in many tasks. In the MRC task, most of the recent research falls into this category. Two main deep learning architectures used by MRC researchers are the *Recurrent Neural Network* (RNN) and *Convolutional Neural Network* (CNN).

RNNs are often used for modeling sequential data by iterating through the sequence elements and maintaining a state containing information relative to what have seen so far. Two common types of RNNs are Long Short-Term Memory (LSTM) (Hochreiter and Schmidhuber 1997) and Gated Recurrent Unit (GRU) (Cho et al. 2014) in unidirectional and bidirectional versions (Chen et al. 2017; Hu, Peng, Huang, et al. 2018; Seo et al. 2017; Kobayashi et al. 2016; Clark and Gardner 2018; Dhingra et al. 2017; Hoang, Wiseman, and Rush 2018; Liu, Zhao, et al. 2018; Chen, Bolton, and Manning 2016; Ghaeini et al. 2018). In MRC systems, like other NLP tasks, these architectures have been commonly used in different parts of the pipeline, such as for representing questions and contexts (Chen et al. 2017; Hu, Peng, Huang, et al. 2018; Seo et al. 2017; Kobayashi et al. 2016; Clark and Gardner 2018; Dhingra et al. 2017; Hoang, Wiseman, and Rush 2018; Liu, Zhao, et al. 2018; Chen, Bolton, and Manning 2016; Ghaeini et al. 2018) or in higher levels of the MRC system such as the modeling layer (Seo et al. 2017; Choi, Hewlett, Uszkoreit, et al. 2017; Li, Li, and Lv 2018). In recent years, the attention-based transformer (Vaswani et al. 2017) has been emerged as a powerful alternative to the RNN architecture. For more detailed information, refer to Section 4.

CNN is a type of deep learning model that is universally used in computer vision applications. It utilizes layers with convolution filters that are applied to local spots of their inputs (LeCun et al. 1998). Originally introduced for computer vision, CNN models have subsequently been shown to be effective for NLP and have achieved excellent results in various NLP tasks (Kim 2014). In MRC systems, CNN is used in the embedding phase (especially, character embedding) (Indurthi et al. 2018; Seo et al. 2017) as well as in the reasoning phase for modeling interactions between the question and passage like in the QANet (Yu et al. 2018). QANet uses CNN and self-attention blocks instead of the RNN, which results in faster answer span detection on the SQuAD dataset (Rajpurkar et al. 2016).

## 3. Input/Output-based analysis

### *3.1 MRC systems input*

The inputs to an MRC system are question and passage texts. The passage is often referred to as context. Moreover, in some systems, the candidate answer list is part of the input.

#### *3.1.1 Question*

Input questions can be grouped into three categories: f*actoid* questions, n*on-factoid* questions, and y*es/no* questions.

Factoid questions are questions that can be answered with simple facts expressed in short text answers like a personal name, temporal expression, or location (Jurafsky and Martin 2019). For example, the answer to the question "Who founded Virgin Airlines?" is a personal name; or questions "What is the average age of the onset of autism?" and "Where is Apple Computer based?" have number and location as an answer, respectively. In other words, the answers to factoid questions are one or more entities or a short expression. Because of its simplicity compared to other types, most research in MRC literature has focused on this type of question (Seo et al. 2017; Chen et al. 2017; Clark and Gardner 2018; Huang et al. 2018).

Non-factoid questions, on the other hand, are open-ended questions that usually require long and complex passage-level answers, such as descriptions, opinions, and explanations (Hashemi et al. 2020). For example, "Why does queen Elizabeth sign her name Elizabeth?", "What is the difference between MRC and QA?", and "What do you think about machine reading comprehension?" are instances of non-factoid questions. In our reviewed papers, 32% of works focus on non-factoid questions. Because of their difficulty, the systems dealing with non-factoid questions have often lower accuracies (Tan, Wei, Yang, et al. 2018; Wang, Yu, Chang, et al. 2018; Wang et al. 2016).

Yes/No questions, as indicated by their name, have yes or no as answers. According to our investigations, the papers which deal with this type of question consider other types of questions as well (Liu, Wei, et al. 2018; Li, Li, and Lv 2018; Zhang et al. 2018).

Refer to Table 2 for the statistics of input/output types in MRC systems. It is clear from the table that the popularity of non-factoid and yes/no questions are increased. Note that since some papers focus on multiple question types, the sum of percentages is greater than 100% in this table.

#### *3.1.2 Context*

The input context can be a *single* passage or *multiple* passages. It is obvious that as the context gets longer, finding the answer becomes harder and more time-consuming. Until now, most of the papers have focused on a single passage (Seo et al. 2017; Yang, Dhingra, et al. 2017; Yang, Hu, et al. 2017; Wang et al. 2017; Zhang et al. 2017). But multiple passages MRC systems are becoming more popular (Huang et al. 2018; Wang, Yu, Guo, et al. 2018; Xie and Xing 2017; Liu, Wei, et al. 2018). According to Table 2, only 4% of the reviewed papers focused on multiple passages in 2016, but this ratio has increased in recent years reaching 52% and 32% in 2019 and 2020, respectively.

### ***3.2 MRC systems output***

The output of MRC systems can be classified into two categories: *abstractive (generative)* output and *extractive (selective)* output.

In the abstractive mode, the answer is not necessarily an exact span in the context and is generated according to the question and context. This output type is especially suitable for non-factoid questions (Choi, Hewlett, Uszkoreit, et al. 2017; Tan, Wei, Yang, et al. 2018; Greco et al. 2016).

In the extractive mode, the answer is a specific span of the context (Seo et al. 2017; Yang, Hu, et al. 2017; Liu, Shen, et al. 2018; Liu et al. 2017; Min, Seo, and Hajishirzi 2017). This output type is appropriate for factoid questions; however, it is possible that the answer to a factoid question may be generative or the answer to a non-factoid question may be extractive. For example, the answer to a non-factoid question may be a whole sentence which is extracted from the context.

There has generally been more focus on extractive MRC systems, but according to Table 2, the popularity of abstractive MRC systems has been increased over recent years. From another point of view, MRC outputs can be categorized as *quiz style, cloze style*, and d*etail style*:

- In the quiz style mode, the answer is one of the multiple candidate answers that must be selected from a predefined set $A$ containing $k$ candidate answers:

$$A = \{a_1, \dots, a_k\},$$

where $a_j$ can be a word, a phrase, or a sentence with length $l_j$:

$$a_j = (a_{j,1}, a_{j,2}, \ldots, a_{j,l_j})$$

- In the cloze style mode, the question includes a blank that must be filled as an answer according to the context. In this case, the answer is often an entity from the context.
- In the detail style mode, there is no candidate or blank, so the answer must be extracted or generated according to the context. In extractive mode, the answer must be a definite range in the context, so it can be shown as $(a_{start}, a_{end})$, where $a_{\text{start}}$ and $a_{end}$ are, respectively, the start and end indices of the answer in the context. In generative mode, on the other hand, the answer is generated from a custom vocabulary $V$.

As shown in Table 2, most of the reviewed papers (72%) have focused on the detail style mode. Also, about 100%, 70%, and 82% of the reviewed papers have focused on factoid questions, multi-passage context, and extractive answers, respectively, due to their lower complexity and existence of rich datasets. For a more detailed categorization of papers based on their input/outputs, refer to Table A1.

**Table2.** Statistics of input/output types in MRC systems.

| | Input | | | | | Output | | | | |
|---|---|---|---|---|---|---|---|---|---|---|
| | Question | | | Context | | | | | | |
| Year | Factoid | Non-factoid | Yes/No | Single passage | Multi- passage | Extractive | Abstractive | Quiz | Cloze | Detail |
| 2016 | 100% | 8% | 0% | 85% | 4% | 77% | 8% | 38% | 54% | 8% |
| 2017 | 100% | 14% | 0% | 100% | 0% | 100% | 9% | 5% | 27% | 95% |
| 2018 | 100% | 26% | 10% | 76% | 34% | 84% | 12% | 18% | 14% | 74% |
| 2019 | 100% | 43% | 2% | 50% | 52% | 76% | 36% | 21% | 3% | 79% |
| 2020 | 100% | 32% | 23% | 68% | 32% | 82% | 50% | 36% | 18% | 59% |
| All | 100% | 30% | 7% | 70% | 42% | 82% | 25% | 21% | 16% | 72% |

## 4. MRC phases

Most of the recent deep learning-based MRC systems have the following phases: *embedding phase*, *reasoning phase*, and *prediction phase*. Many of the reviewed papers focus on developing new structures for these phases, especially the reasoning phase.

### *4.1 Embedding phase*

In this phase, input characters, words, or sentences are represented by real-valued dense vectors in a meaningful space. The goal of this phase is to provide question and context embedding. Different levels of embedding are used in MRC systems. Character-level and word-level embeddings can capture the properties of words, and higher level representations can represent syntactic and semantic information of input text. Table 3 shows the statistics of various embedding methods used in the reviewed papers. Since there is not any paper that uses the character embedding as the only embedding method, there is no character embedding column in this table. For a complete list of papers categorized based on their embedding methods, refer to Table A2.

**Table 3.** Statistics of different embedding methods used by reviewed papers.

| Year | Word embedding | Hybrid (word-char embedding) | Sentence embedding |
|---|---|---|---|
| 2016 | 100% | 0% | 5% |
| 2017 | 56% | 40% | 4% |
| 2018 | 45% | 54% | 6% |
| 2019 | 67% | 33% | 3% |
| 2020 | 86% | 14% | 18% |
| All | 64% | 35% | 7% |

#### *4.1.1 Character embedding*

Some papers use character embedding as part of their embedding phase. This type of embedding is useful to overcome unknown and rare words problems (Dhingra et al. 2016; Seo et al. 2017). To generate the input representation, deep neural network models are commonly used. Inspired by Kim's work (2014), some papers have used CNN models to embed the input characters (Seo et al. 2017; Zhang et al. 2017; Gong and Bowman 2018; Kundu and Ng 2018a). Some other papers have used character level information captured from the final state of an RNN model like LSTM (or BiLSTM) and GRU (or BiGRU) (Yang, Dhingra, et al. 2017; Du and Cardie 2018; Hu, Peng, Huang, et al. 2018; Wang, Yuan, and Trischler 2017; Wang, Yu, Guo, et al. 2018). As another approach which uses both CNN and LSTM to embed input characters, LSTM-char CNN (Kim et al. 2016) is also used in MRC literature (Prakash, Tripathy, and Banu 2018). We classify these papers in two categories, CNN and RNN, and so the sum of percentages is greater than 100% in Figure 2.

Figure 2 shows the percentage of different character embedding methods over different years. Other methods include skip-gram, n-grams, and more recent methods like ELMo (Peters et al. 2018). The overall trend shows a relative decrease in the usage of RNN-based methods and a relative increase in the usage of CNN-based methods.

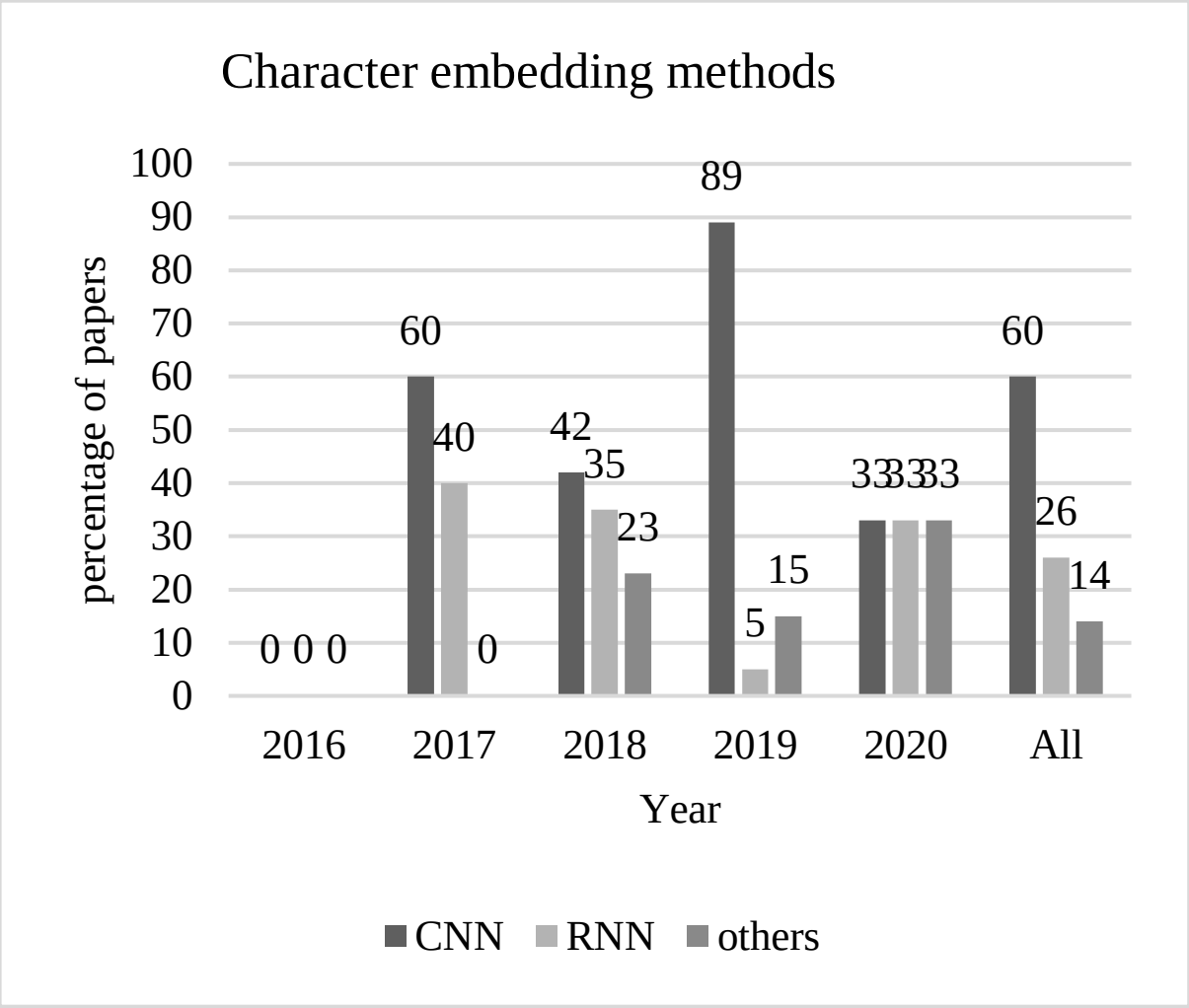

**Figure 2**. The percentage of different character embedding methods over different years

*4.1.2 Word embedding*

Word embedding is to represent the words in a numeric vector space, which is performed by two main approaches: 1) non-contextual embedding, and 2) contextual embedding.

**Non-contextual word embedding**

Non-contextual embeddings present a single general representation for each word, regardless of its context. There are three main non-contextual embeddings: 1) One-hot encoding, 2) learning word embedding jointly with the main task, and 3) using pre-trained word embeddings (fixed or fine-tuned).

*Note that some works use multiple methods, so the sum of percentages in the tables may be greater than 100%.*

One-hot encoding is the most basic way to turn a token into a vector. These are binary, sparse, and very high dimensional vectors with the size of the vocabulary (the number of unique words in the corpus). To represent a word *w*, all the vector elements are set to zero, except the one which identifies *w*. This approach has been less popular than other approaches in recent papers (Cui et al. 2016; Liu and Perez 2017).

Another popular way to represent words is learned word embedding, which delivers dense real-valued representations. In the presence of a large amount of training data, it is advised to learn the word embeddings from scratch jointly with the main task (Chollet 2017).

Some studies have shown that initializing word embeddings with pre-trained values results in better accuracies than random initialization (Dhingra et al. 2017; Wang, Zhang, et al. 2020; Zhou, Luo, and Wu 2020; Ren, Cheng, and Su 2020). This approach is especially useful in the low-data scenarios (Chollet 2017; Dhingra et al. 2017). GloVe embedding (Pennington, Socher, and Manning 2014) is most common pre-trained word representation among non-contextual representations, used in the reviewed papers (Xiong, Zhong, and Socher 2017; Yin, Ebert, and Schütze 2016; Chen, Bolton, and Manning 2016; Chen et al. 2017; Liu et al. 2017; Gong and Bowman 2018; Wang, Yuan, and Trischler 2017; Wang, Yu, Guo, et al. 2018). Word2Vec (Mikolov et al. 2013) is another word embedding used in this task (Kobayashi et al. 2016; Šuster and Daelemans 2018; Chaturvedi, Pandit, and Garain 2018). These pre-trained word embeddings are fine-tuned (Chen, Bolton, and Manning 2016; Kobayashi et al. 2016; Clark and Gardner 2018; Šuster and Daelemans 2018; Liu, Shen, et al. 2018; Zhang et al. 2017) or left as fixed embeddings (Seo et al. 2017; Weissenborn, Wiese, and Seiffe 2017; Shen et al. 2017; Gong and Bowman 2018). Fine-tuning some keywords such as "what", "how", "which", and "many" could be crucial for QA systems, while most of the pre-trained word embeddings can be kept fixed (Chen et al. 2017).

Finally, it is worth noting that some papers use hand-designed word features such as named entity (NE) tag and part-of-speech (POS) tag along with embedding of words (Huang et al. 2018). Table 4 shows the statistics of these approaches through different years in the reviewed papers.

**Table 4**. Statistics of different word representation methods in the reviewed papers.

| | NON-CONTEXTUAL | | | | CONTEXTUAL | | | |
|---|---|---|---|---|---|---|---|---|
| YEAR | ONE HOT ENCODING | LEARNED WORD EMBEDDING | FIXED PRE-TRAINED | FINE-TUNED | LEARNED WORD EMBEDDING | | FIXED PRE-TRAINED | FINE-TUNED |
| | | | | | RNN | CNN | | |
| 2016 | 11% | 56% | 22% | 22% | 67% | 11% | 0% | 0% |
| 2017 | 4% | 12% | 36% | 40% | 41% | 0% | 0% | 0% |
| 2018 | 4% | 8% | 29% | 18% | 33% | 6% | 4% | 0% |
| 2019 | 0% | 0% | 38% | 21% | 0% | 0% | 5% | 45% |
| 2020 | 9% | 14% | 27% | 14% | 50% | 0% | 23% | 36% |
| All | 4% | 9% | 38% | 27% | 46% | 2% | 4% | 21% |

**Contextual word embedding**

Despite the relative success of non-contextual embeddings, they are static, so all meanings of a word are represented with a fixed vector (Ethayarajh 2019). Different from static word embeddings, contextual embeddings move beyond word-level semantics and represent each word considering its context (surrounding words). To obtain the context-based representation of the words, two approaches can be adopted: 1) learning the embeddings jointly with the main task, and 2) using pre-trained contextual embeddings (fixed or fine-tuned).

For learning the contextual word embedding, a sequence modeling method, usually an RNN, is used. For example, Chen et al. (Chen et al. 2017) used a multi-layer BiLSTM model for this purpose. In Yang, Kang, and Seo (2020) study, forward and backward GRU hidden states are combined to generate contextual representations of query and context words. Bajgar, Kadlec, and Kleindienst (2017) used different approaches for the query and context words, where the combination of forward and backward GRU hidden states are exploited for representing the context words, while the final hidden state of GRU is used for the query. On the other hand, pre-trained contextualized embeddings such as ELMo (Peters et al. 2018), BERT (Devlin et al. 2018), and GPT (Radford et al. 2018) are deep neural language models that are trained on large unlabeled corpuses. The ELMo method (Peters et al. 2018) obtains the contextualized embeddings by a 2-layer Bi-LSTM, while BERT (Devlin et al. 2018) and GPT (Radford et al. 2018) are bi-directional and uni-directional transformer-based (Vaswani et al. 2017) language models, respectively. These embeddings are used either besides other embeddings (Hu, Peng, Huang, et al. 2018; Hu, Peng, Wei, et al. 2018; Seo et al. 2018; Lee and Kim 2020; Ren, Cheng, and Su 2020) or alone (Bauer, Wang, and Bansal 2018; Zheng et al. 2020).

Due to the success of the contextual word embeddings in many NLP tasks, there is a clear trend toward using these embeddings in recent years (Wang and Bansal 2018; Hu, Peng, Huang, et al. 2018; Hu, Peng, Wei, et al. 2018; Bauer, Wang, and Bansal 2018). This is obvious in Table 4, where the use of fixed pre-trained and fine-tuned contextual embeddings have increased from 0% in 2016 to 23% and 36% in 2020, respectively. Note that some papers use multiple methods, so the sum of percentages in the tables may be greater than 100%.

#### *4.1.3 Hybrid word-character embedding*

The combination of word embedding and character embedding is used in some reviewed papers (Zhang et al. 2017; Gong and Bowman 2018; Seo et al. 2017; Yang, Dhingra, et al. 2017). Hybrid embedding tries to use the strengths of both word and character embeddings. A simple approach is to concatenate the word and character embeddings. As an example, (Lee and Kim 2020) used GloVe as the word embedding and the output of the CNN model as the character embedding.

This approach suffers from a potential problem. Word embedding has better performance for frequent words, while it can have negative effects for representing rare words. The reverse is true for character embedding (Yang, Dhingra, et al. 2017). To solve this problem, some researchers introduced a gating mechanism which regulates the flow of information. Yang, Dhingra, et al. (2017) used a fine-grained gating mechanism for dynamic concatenation of word and characters embedding. This mechanism uses a gate vector, which is a linear multiplication of word features (POS and NE), to control the flow of information of word and character embeddings. Seo et al. (2017) used highway networks (Srivastava, Greff, and Schmidhuber 2015) for embedding concatenation. These networks use the gating mechanism learned by the LSTM network. According to Table 3, the use of hybrid embedding in reviewed papers has increased from 0% in 2016 to 54% in 2018. However, with the success of language model-based contextual embeddings, the direct combination of character and word embeddings has decreased thereafter.

#### *4.1.4 Sentence embedding*

Sentence embedding is a high-level representation in which the entire sentence is encoded in a single vector. It is often used along with other embeddings (Yin, Ebert, and Schütze 2016). However, sentence embedding is not so popular in MRC systems, because the answer is often a sentence part, not the whole sentence.

### ***4.2 Reasoning phase***

The goal of this phase is to match the input query (question) with the input document (context). In other words, this phase determines the related parts of the context for answering the question by calculating the relevance between question and context parts. Recently, Phrase Indexed Question Answering (PIQA) model (Seo et al. 2018) enforces complete independence between document encoder and question encoder and does not include any cross attention between question and document. In this model, each document is processed beforehand, and its *phrase index vectors* are generated. Then, at inference time, the answer is obtained by retrieving the nearest indexed phrase vector to the query vector.

The attention mechanism (Bahdanau, Cho, and Bengio 2015), originally introduced for machine translation, is used for this phase. In recent years, with the advent of attention-based transformer architecture (Vaswani et al. 2017) as an aleternative to common sequential structures like RNN, new transformer-based language models, such as BERT (Devlin et al. 2018) and XLNet (Yang, Dai, et al. 2019), have been introduced. They are used as the basis for new state-of-the-art results in MRC task (Sharma and Roychowdhury 2019; Yang, Wang, et al. 2019; Su et al. 2019; Zhang, Zhao, et al. 2020; Zhang, Luo, et al. 2020; Tu et al. 2020) by adding or modifying final layers and fine-tuning them on the target task.

The attention mechanism used in MRC systems can be explored in three perspectives: *direction*, *dimension*, and *number of steps.* For the statistics, refer to Table 5.

#### *4.2.1 Direction*

Some research only use the context-to-query (C2Q) attention vector (Wang and Jiang 2017; Cui et al. 2016; Weissenborn, Wiese, and Seiffe 2017; Huang et al. 2018) called *one directional* attention mechanism. It signifies which query words are relevant to each context word (Cui, Chen, et al. 2017; Seo et al. 2017).

In *bi-directional* attention mechanism, query-to-context (Q2C) attention weights are also calculated (Cui, Chen, et al. 2017; Seo et al. 2017; Xiong, Zhong, and Socher 2017; Clark and Gardner 2018; Min, Seo, and Hajishirzi 2017; Liu et al. 2017) along with C2Q. It signifies which context words have the closest similarity to one of the query words and are hence critical for answering the question (Cui, Chen, et al. 2017; Seo et al. 2017). In transformer-based MRC models like BERT-based models, the question and context are processed as one sequence, so the attention mechanism can be considered as bi-directional attention. As shown in Table 5, the ratio of bi-directional attention usage has increased in recent years.

#### *4.2.2 Dimension*

There are two attention dimensions in the reviewed papers: *one-dimensional* and *two-dimensional* attentions. In one-dimensional attention, the whole question is represented by one embedding vector, which is usually the last hidden state of the contextual embedding (Chen, Bolton, and Manning 2016; Kadlec et al. 2016; Dhingra et al. 2017; Weissenborn, Wiese, and Seiffe 2017; Shen et al. 2017). It does not pay more attention to important question words. On the contrary, in two-dimensional attention, every word in the query has its own embedding vector (Cui, Chen, et al. 2017; Seo et al. 2017; Chen et al. 2017; Yang, Hu, et al. 2017; Clark and Gardner 2018).

According to Table 5, 86% of all reviewed papers use two-dimensional attention. Also, the use of two-dimensional attention has increased over recent years.

#### *4.2.3 Number of steps*

According to the number of reasoning steps, three types of MRC systems can be seen: s*ingle-step reasoning*, *multi-step reasoning with a fixed number of steps,* and *dynamic multi-step reasoning*.

In the single step reasoning, question and passage matching is done in a single step. However, the obtained representation can be processed through multiple layers to extract or generate the answer (Seo et al. 2017; Chen, Bolton, and Manning 2016; Clark and Gardner 2018). In multi-step reasoning, question and passage matching is done in multiple steps such that the question-aware context representation is updated by integrating the intermediate information in each step. The number of steps can be static (Yang, Hu, et al. 2017; Hu, Peng, Huang, et al. 2018) or dynamic (Shen et al. 2017; Song et al. 2018; Dhingra et al. 2017). Dynamic multi-step reasoning uses a termination module to decide whether the inferred information is sufficient for answering or more reasoning steps are still needed. Therefore, the number of reasoning steps in this model depends on the complexity of the passage and question. It's obvious that in multi-step reasoning, the model complexity increases by the number of reasoning steps. In the transformer-based MRC models, the number of steps is fixed and depends on to the number of layers.

According to Table 5, about 61% of reviewed papers use single step reasoning, but the popularity of multi-step reasoning has increased over recent years. For a detailed list of the used reasoning methods in different papers refer to Table A3.

**Table 5.** Statistics of different attention mechanisms used in the reasoning phase of MRC systems.

| | Direction | | Dimension | | Number of steps | | |
|---|---|---|---|---|---|---|---|
| Year | One-direction | Two-direction | One-dimension | Two- dimension | Single | Multi- fixed | Multi-dynamic |
| 2016 | 89% | 11% | 78% | 22% | 89% | 22% | 0% |
| 2017 | 52% | 38% | 10% | 90% | 71% | 10% | 5% |
| 2018 | 51% | 49% | 21% | 83% | 74% | 21% | 2% |
| 2019 | 12% | 89% | 2% | 98% | 44% | 56% | 4% |
| 2020 | 18% | 77% | 18% | 82% | 55% | 46% | 0% |
| All | 35% | 64% | 15% | 86% | 61% | 36% | 3% |

### *4.3 Prediction phase*

The final output of an MRC system is specified in the prediction phase. The output can be extracted from context or generated according to context. In generation mode, a decoder module generates answer words one by one (Hewlett, Jones, and Lacoste 2017). In some cases, multiple choices are presented to the system, and it must select the best answer according to the question and passage(s) (Greco et al. 2016). These multi-choice systems can be seen in both extractive and generative models based on whether the answer choices occur in the passage or not.

The extraction mode is implemented in different forms. If the answer is a span of context, the start and end indices of the span are predicted in many studies by estimating the probability distribution of indices over the entire context (Chen et al. 2017; Yang, Hu, et al. 2017; Clark and Gardner 2018; Yang, Dhingra, et al. 2017; Wang and Jiang 2017; Xiong, Zhong, and Socher 2017).

In some studies, the candidate chunks (answers) are extracted first, which are ranked by a trained model. These chunks can be sentences (Min, Seo, and Hajishirzi 2017; Duan et al. 2017) or entities (Sachan and Xing 2018). In Ren, Cheng, and Su (2020) study, after extracting the candidate chunks from various contexts, a linear transformation is used along with a sigmoid function to compute the score of the answer candidates.

Table 6 shows the statistics of these categories in the reviewed papers. It is clear that most papers (65%) extract the answer span in the passage(s). It seems that developing rich span-based datasets like SQuAD (Rajpurkar et al. 2016) is the reason for this popularity. Also, the generation-based papers have increased from 10% in 2016 to 55% in 2020. For more details, refer to Table A4.

**Table 6.** Statistics of different prediction phase categories in the reviewed papers.

| YEAR | EXTRACTION MODE | | GENERATION MODE | |
|---|---|---|---|---|
| | SPAN DETECTION | CANDIDATE RANKING | ANSWER GENERATION | CANDIDATE RANKING |
| 2016 | 40% | 50% | 0% | 10% |
| 2017 | 70% | 26% | 4% | 0% |
| 2018 | 66% | 20% | 8% | 8% |
| 2019 | 69% | 5% | 16% | 19% |
| 2020 | 59% | 23% | 32% | 23% |
| All | 65% | 18% | 13% | 13% |

## 5. MRC datasets

Rich datasets are the first prerequisite for having accurate machine learning models. Especially, deep neural network models require high volumes of training data to achieve good results. For this reason, in recent years, many researchers have focused on collecting big datasets. For example, Stanford Question Answering Dataset (SQuAD) (Rajpurkar et al. 2016), which is a popular MRC dataset used in many studies, includes over 100,000 training samples.

MRC datasets can be categorized according to their volume, domain, question type, answer type, context type, data collection method, and language.

In terms of domain, MRC datasets can be classified into two categories: *open domain* and *close domain*. Open domain datasets contain diverse subjects, while close domain datasets focus on specific areas such as the medical domain. For example, the SQuAD (Rajpurkar et al. 2016) dataset, which contains Wikipedia articles, is an open domain dataset and emrQA (Pampari et al. 2018),

BIOMRC (Pappas et al. 2020), LUPARCQ (Horbach et al. 2020) is a close domain dataset with biology as its subject.

There are two data collection approaches in MRC datasets, *automatic* approach, and *crowdsourcing* approach. The former generates questions/answers without direct human interventions. For instance, datasets that contain cloze-style questions, such as Children's Book Test dataset (Hill et al. 2016), are generated by removing important entities from text. Also, in some datasets, questions are automatically extracted from the search engine's user logs (Nguyen et al. 2016) or real reading comprehension tests (Lai et al. 2017).

On the other hand, in the crowdsourcing approach, humans generate questions, answers, or select related paragraphs. Of course, a dataset can be generated by a combination of these two approaches. For instance, in MS MARCO (Nguyen et al. 2016), questions have been generated automatically, while these questions have answered and evaluated by crowdsourcing.

Table 7 shows a detailed list of the datasets proposed from 2016 to 2020. In this table, the datasets with a link address are publicly available.

**Table 7**. MRC datasets proposed from 2016 to 2020. (A: Answer, P: passage, Q: Question)

| Dataset | Open/Close Domain | Language | Question Type | Context Type | Answer Type | #Question | #Context | Collect Data | Question Classification | Link Address |
|---|---|---|---|---|---|---|---|---|---|---|
| MS MARCO (Nguyen et al. 2016) | Open | English | Factoid | Multi-document | Abstractive | 100K | 1M Passage +200K Document | Q: Automatic A:Crowdsourced | Yes | http://www.msmarco.org |
| Newsqa (Trischler et al. 2017) | Open | English | Factoid | Single paragraph | Extractive (Detail) | 100K | 10K Articles | Crowdsourced | No | https://www.microsoft.com/en-us/research/project/newsqa-dataset/ |
| BookTest (NE, CN) (Bajgar, Kadlec, and Kleindienst 2017) | Open | English | Factoid | Single paragraph | Extractive (Cloze Style) | 14M | 13.5K Books | Automatic | No | - |
| People Daily news dataset (Cui et al. 2016) | Open | Chinese | Factoid | Single paragraph | Extractive (Cloze Style) | 876K | 60k Articles | Automatic | No | - |
| Children's Fairy Tale (CFT) (Cui et al. 2016) | Open | Chinese | Factoid | Single paragraph | Extractive (Cloze Style) | 3.5K | 60 k Passages | Automatic | No | - |
| SQuAD (Rajpurkar et al. 2016) | Open | English | Factoid | Single paragraph | Extractive (Detail) | 100K | 536 Articles | Crowdsourced | No | https://stanford-qa.com |
| Who did what (Onishi et al. 2016) | Open | English | Factoid | Single paragraph | Extractive (Quiz Style) | 330K | 200 k Passages | Automatic | No | https://tticnlp.github.io/who_did_what/ |
| CliCl (Šuster and Daelemans 2018) | Close (medical) | English | Factoid | Single paragraph | Extractive (Cloze Style) | 105K | 12K Passages | Automatic | No | https://github.com/clips/clicr |
| DRCD (Shao et al. 2018) | Open | Chinese | Factoid | Single paragraph | Extractive (Detail) | 30K | 10K Paragraphs from 2K articles | Crowdsourced | Yes | https://github.com/DRCSolutionService/DRCD |
| DuoRC (Saha et al. 2018) | Open | English | Factoid and non-Factoid | Multi-paragraph | Abstractive | 186K | 7.5K Passages | Crowdsourced | No | https://duorc.github.io/ |
| QBLink (Elgohary, Zhao, and Boyd-Graber 2018) | Open | English | Factoid | Multi-paragraph | Extractive (Detail) | 56K | Context is extracted before reading. | Automatic | No | https://sites.google.com/view/qanta/projects/qblink |
| SQuAD-T (Tan, Wei, Zhou, et al. 2018) | Open | English | Factoid | Single Paragraph | Extractive (Detail) | 100K | 536 Articles | Crowdsourced | No | https://github.com/Chuanqi1992/SQuAD-T |
| SQuAD 2.0 (Rajpurkar, Jia, and Liang 2018) | Open | English | Factoid | Single paragraph | Extractive (Detail) | 150K | 505 Articles | Crowdsourced | No | https://rajpurkar.github.io/SQuAD-explorer |
| TriviaQA (Joshi et al. 2017) | Open | English | Factoid | Multi-document | Extractive (Detail) | 95K | 650k Passages | Automatic | No | http://nlp.cs.washington.edu/triviaqa/ |
| Race (Lai et al. 2017) | Open | English | Factoid and non-Factoid | Multi-paragraph | Abstractive (Quiz style) | 97K | 27K Passages | Automatic | No | https://www.cs.cmu.edu/~glai1/data/race |
| DuReader (He et al. 2018) | Open | Chinese | Factoid and non-Factoid | Multi-document | Abstractive | 200K Questions, 420K answers | 1M Passages | Q & P: Automatic<br>A: Crowdsourced | Yes | http://ai.baidu.com/broad/download?dataset=dureader |
| SciQ (Welbl, Liu, and Gardner 2017) | Close (science) | English | Factoid | Single document | Extractive (Quiz style) | 13.7K | 13.7K Passages | Crowdsourced | No | https://allenai.org/data/sciq |
| QuAC (Choi et al. 2018) | Open (dialog) | English | Factoid and non-Factoid | Single paragraph | Extractive (Detail-no-answer) | 100K | 14K Passages | Crowdsourced | No | http://quac.ai |
| CLOTH (Xie et al. 2018) | Open | English | Factoid | Single paragraph | Abstractive (Quiz style) | 99K | 7K Passage | Automatic | Yes | - |
| emrQA (Pampari et al. 2018) | Close (electronic medical records) | English | Factoid | Single paragraph | Extractive (Detail) | 455K | 2K Passages | Automatic | No | https://www.i2b2.org/NLP/DataSets/ |
| MultiRC (Khashabi et al. 2018) | Open | English | Factoid and non-Factoid | Single paragraph | Abstractive (Quiz style) | 6K | +800 Passages | Crowdsourced | No | http://cogcomp.org/multirc/ |
| MCScript (Ostermann et al. 2018) | Open | English | Factoid and non-Factoid | Single paragraph | Abstractive | 32K | 2K Passages | Crowdsourced | No | http://www.sfb1102.uni-saarland.de/?page_id=2582 |
| Children's Book Test (Hill et al. 2016) | Open | English | Factoid | Single paragraph | Extractive (Cloze Style) | 687343 | 108 book | Automatic | No | https://research.fb.com/downloads/babi/ |
| NarrativeQA (Kočiský et al. 2018) | Open | English | Factoid and non-Factoid | Multi-paragraph | Abstractive | 46,765 | 1,572 stories (books, movie scripts) | Crowdsourced | No | https://github.com/deepmind/narrativeqa |
| Natural Question (Kwiatkowski et al. 2019) | Open | English | Factoid and non-Factoid | Multi-paragraph | Abstractive | 323045 | - | Q:Automatic<br>A:Crowdsourced | No | https://ai.google.com/research/NaturalQuestions |
| Wikihop (Welbl, Stenetorp, and Riedel 2018) | Open | English | Factoid | Multi-paragraph | Extractive(Quiz Style) | 51318 | 3-63 for each Q | Q: Automatic<br>A: Crowdsourced | No | http://qangaroo.cs.ucl.ac.uk/ |
| Medhop (Welbl, Stenetorp, and Riedel 2018) | Close (molecular biology) | English | Factoid | Multi-paragraph | Extractive(Quiz Style) | 2508 | 5-64 for each Q | Q: Automatic<br>A: Crowdsourced | No | http://qangaroo.cs.ucl.ac.uk/ |
| HotpotQA (Yang et al. 2018) | Open | English | Factoid | Multi-paragraph | Extractive (Detail, yes/no) | 112779 | - | Crowdsourced | No | https://hotpotqa.github.io/ |
| RACE-C (Liang, Li, and Yin 2019) | Open | English | Factoid and non-Factoid | Multi-paragraph | Abstractive (Quiz style) | 14K | 4K | Automatic | No | - |
| CMRC 2018 (Cui, Liu, et al. 2019) | Open | Chinese | Factoid and non-Factoid | Single paragraph | Extractive (Detail) | 20K | - | Crowdsourced | No | https://bit.ly/2ZdS8Ct |
| AmazonQA (Gupta et al. 2019) | Open | English | Factoid and non-Factoid | Multi-paragraph | Extractive (Detail) | 923k | 14M | Crowdsourced | Yes | https://github.com/amazonqa/amazonqa |
| (Hardalov, Koychev, and Nakov 2019) | Close (history, biology, geography and philosophy) | Bulgarian | Factoid and non-Factoid | - | Abstractive (Quiz style) | 2.6K | - | Automatic | Yes | https://github.com/mhardalov/bg-reason-BERT |
| BiPaR (Jing, Xiong, and Zhen 2019) | Close (novels) | Chinese English | Factoid and non-Factoid | Single paragraph | Extractive (Detail) | 14.7K | 3.7K | Crowdsourced | No | https://multinlp.github.io/BiPaR/ |
| COSMOS QA (Huang, Le Bras, et al. 2019) | Open | English | Factoid and non-Factoid | Single paragraph | Abstractive (Quiz style) | 35.6K | 21.9K | Crowdsourced | No | https://wilburone.github.io/cosmos |
| DROP (Dua et al. 2019) | Open | English | Factoid | Single paragraph | Extractive and Abstractive | 96.6K | 6.6K | Crowdsourced | No | https://allennlp.org/drop |
| FaQuAD (Sayama, Araujo, and Fernandes 2019) | Close (higher education) | Brazilian | Factoid | Single paragraph | Extractive | 900 | 249 | Crowdsourced | No | https://github.com/liafacom/faquad |
| MCScript2.0 (Ostermann, Roth, and Pinkal 2019) | Close (narrative) | English | Factoid | Single paragraph | Abstractive (Quiz style) | 20K | 3.5K | Crowdsourced | No | http://www.sfb1102.uni-saarland.de/?page_id=2582 |
| ARCD (Mozannar et al. 2019) | Open | Arabic | Factoid | Single paragraph | Extractive (Detail) | 1.4K | 155 Article | Crowdsourced | No | https://github.com/husseinmozannar/SOQAL |
| QUOREF (Dasigi et al. 2019) | Open | English | Factoid | Single paragraph | Extractive (Detail) | 24K | 4.7K | Crowdsourced | No | https://allennlp.org/quoref |
| SMART (Yao et al. 2019) | Open | Chinese | Factoid | Single paragraph | Extractive (Detail) | 39.4K | 564 Article | Crowdsourced | Yes | - |
| XQA (Liu, Lin, et al. 2019) | Open | 9 languages | Factoid | Multi-paragraph | Extractive (Detail) | 90k | top-10 Wikipedia articles for each question | Automatic | No | http://github.com/thunlp/XQA |
| HindiRC (Anuranjana, Rao, and Mamidi 2019) | Open | Hindi | Factoid and non-Factoid | Single paragraph | Extractive (Detail) | 127 | 24 | Crowdsourced | Yes | https://github.com/erzaliator/HindiRC-Data |
| ROPES (Lin et al. 2019) | Open | English | Factoid | Single paragraph | Extractive (Detail) | 14.3K | 1.4K | Crowdsourced | No | https://allennlp.org/ropes |
| (Li, Li, and Liu 2019) | Open | Chinese | Factoid and non-Factoid | Single paragraph | Extractive (Detail) | 242K | 85K | Crowdsourced | No | - |
| BIOMRC (Large) (Pappas et al. 2020) | Close (Biomedical) | English | Factoid | Single Paragraph | Extractive (Cloze Style) | 812707 | 812707 | Automatic | No | https://archive.org/details/biomrc_dataset |
| (Watarai and Tsuchiya 2020) | Open | Japanese | Factoid | Single Paragraph | Extractive (Quiz and Cloze Style) | 22790 | 64040 | Automatic | No | - |
| C$^3$(Sun et al. 2020) | Open | Chinese | Factoid | Multi-paragraph | Extractive (Quiz Style) | 19577 | 13369 | Automatic | No | https://dataset.org/c3/ |
| R$^4$C(Inoue, Stenetorp, and Inui 2020) | Open | English | Factoid | Multi-paragraph | Extractive (Detail) | 4588 | 45880 (10 Paragraphs per Question) | Crowdsourced | No | https://naoya-i.github.io/r4c/ |
| ReClor (Yu et al. 2020) | Open | English | non-Factoid | Single paragraph | Abstractive (Quiz Style) | 6138 | 6138 | Automatic | No | https://whyu.me/reclor |
| ReCo (Wang, Yao, et al. 2020) | Open | Chinese | Factoid | Single paragraph | Generative (Quiz Style) | 300K | 300K | Q: Automatic P, A: Crowdsourced | No | https://github.com/benywon/ReCO |
| SQuAD2-CR (Lee, Hwang, and Cho 2020) | Open | English | Factoid | Single paragraph | Extractive (Detail) | 150K (16403 annotated) | 500 article | Crowdsourced | No | https://antest1.github.io/SQuAD2-CR/ |
| OneStepQA (Berzak, Malmaud, and Levy 2020) | Open | English | non-Factoid | Single paragraph | Abstractive (Quiz Style) | 1458 (486 × three version) | 30 article (162 paragraph) | Crowdsourced | No | https://github.com/berzak/onestop-qa |
| LUPARCQ (Horbach et al. 2020) | Close (Biology) | English(E), German(G), Basque(B) | Factoid | Multi-paragraph | Extractive (Detail) | E: 299<br>G:95<br>E: 152 | E: 24<br>G:21<br>B:18 | Crowdsourced, Automatic | No | - |

## 6. MRC evaluation measures

Based on the system output type, different evaluation metrics are introduced. We classify these measures into two categories: *extractive* metrics and g*enerative* metrics.

### 6.1 Extractive metrics

These metrics are used for the extractive outputs. Table 8 shows the statistics of these measures in the reviewed papers.

- ***F1 score:*** The harmonic mean of precision and recall is a common extractive metric for evaluating MRC systems. It takes into account the system output and the ground-truth answer as bag-of-tokens (words). Precision is calculated as the number of correctly predicted tokens divided by the number of all predicted tokens. The recall is also the number of correctly predicted tokens divided by the number of ground- truth tokens. The F1 score is then calculated as:

$$F1 = 2 \times \frac{precision \times recall}{precision + recall} \qquad (1)$$

The final F1 score is then obtained by averaging over all question-answer pairs.

- ***Exact Match (EM)***. This is the percentage of answers that exactly match with the correct answers. If there are multiple answers to a question in a dataset, a match with at least one of the answers is considered as an exact match. Some QA systems such as multiple-choice QA systems (Zhang, Wu, et al. 2020) or sentence selection QA systems (Min, Seo, and Hajishirzi 2017) call this measure as accuracy (ACC) instead of EM.

- ***Mean Average Precision (MAP).*** This measure is used when the system returns several answers along with their ratings. The MAP for a set of question-answer pairs is the mean of Average Precision scores (AveP) for each one.

$$MAP = \frac{\sum_{q=1}^{Q} AveP(q)}{Q}, \qquad (2)$$

where Q is the number of queries. AveP is an evaluation measure used in information retrieval systems. It evaluates a ranked list of documents in response to a given query. In MRC literature, the ranked list of answers for a given query is evaluated. AveP is computed as the average of precisions over the interval from recall=0 to recall=1 in the precision-recall curve (Turpin and Scholer 2006).

- ***Mean Reciprocal Rank (MRR).*** This is a common evaluation metric for factoid QA systems introduced in TREC QA track 1999. According to the definition presented in the “Evaluation of Factoid Answers” Section of the “Speech and Language Processing” book (Jurafsky and Martin 2019), MRR evaluates a ranked list of answers based on the inverse of the rank of the correct answer. For example, if the rank of the correct answer in the output list of a system is 4, the reciprocal rank score for that question would be 1/4. This measure is then averaged for all questions in the test set.

- ***Precision@K.*** This measure is also borrowed from information retrieval literature. It is the number of correct answers in the first k returned answers without considering the position of these correct ones (Manning, Raghavan, and Schütze 2008).

- ***Hit@K or Top-K.*** Hit@K, which is equivalent to the Top-K accuracy, counts the number of samples where their first k returned answers include the correct answer.

Table 8. Statistics of evaluation measures used in reviewed papers

| Year | Extractive Metrics | | | | | | | | Generative Metrics | | | |
|---|---|---|---|---|---|---|---|---|---|---|---|---|
| | EM | F1 | MAP | MRR | P@1 | R@1 | ACC | Hit@k/ Top-k | ROUGE_L | BLEU | METEOR | CIDEr |
| 2016 | 0% | 0% | 13% | 13% | 0% | 0% | 87% | 13% | 5% | 0% | 0% | 0% |
| 2017 | 59% | 63% | 7% | 7% | 4% | 0% | 41% | 0% | 3% | 7% | 0% | 3% |
| 2018 | 58% | 80% | 4% | 2% | 7% | 7% | 36% | 0% | 22% | 18% | 7% | 2% |
| 2019 | 64% | 64% | 2% | 2% | 3% | 2% | 2% | 0% | 19% | 17% | 5% | 0% |
| 2020 | 46% | 55% | 0% | 0% | 23% | 18% | 50% | 0% | 9% | 14% | 0% | 0% |
| All | 56% | 60% | 4% | 3% | 7% | 5% | 29% | 1% | 16% | 14% | 4% | 1% |

### *6.2 Generative metrics*

The metrics used for evaluating the performance of generative MRC systems are the same metrics used for machine translation and summarization evaluation. Table 9 shows the statistics of these measures in the reviewed papers.

- ***Recall-Oriented Understudy for Gisting Evaluation (ROUGE).*** This measure compares a system-generated answer with the human-generated one (Lin 2004). It is defined as the recall of the system based on the n-grams, i.e., the number of correctly generated n-grams divided by the total number of n-grams in the human-generated answer.

- ***BiLingual Evaluation Understudy (BLEU).*** This metric first introduced for evaluating the output of machine translation task. It is defined as the precision of the system based on the n-grams, i.e., the number of correctly generated n-grams divided by the total number of n-grams in the system-generated answer (Papineni et al. 2002).

- ***Metric for Evaluation of Translation with Explicit Ordering (METEOR).*** This measure is designed to fix some weaknesses of the popular BLEU measure. METEOR is based on an alignment between the system output and reference output. It also introduces a penalty to have longer matches between two strings (Banerjee and Lavie 2005).

- ***Consensus-based Image Description Evaluation (CIDEr).*** This measure is initially introduced for evaluating the image description generation task (Vedantam, Lawrence Zitnick, and Parikh 2015). It is based on the n-gram matching of the system output and reference output in the stem or root forms. According to this measure, the n-grams that are not in the reference output should not be in the system output. Also, the common n-grams in the dataset are less informative and have lower weights.

Figure 3 shows the ratio of the used extractive/generative measures in the reviewed papers. According to this figure, the generative measures have been more popular in recent years than in 2016 and 2017. The obvious reason for this is the trend toward developing abstractive MRC systems. For more details, refer to Table A5.

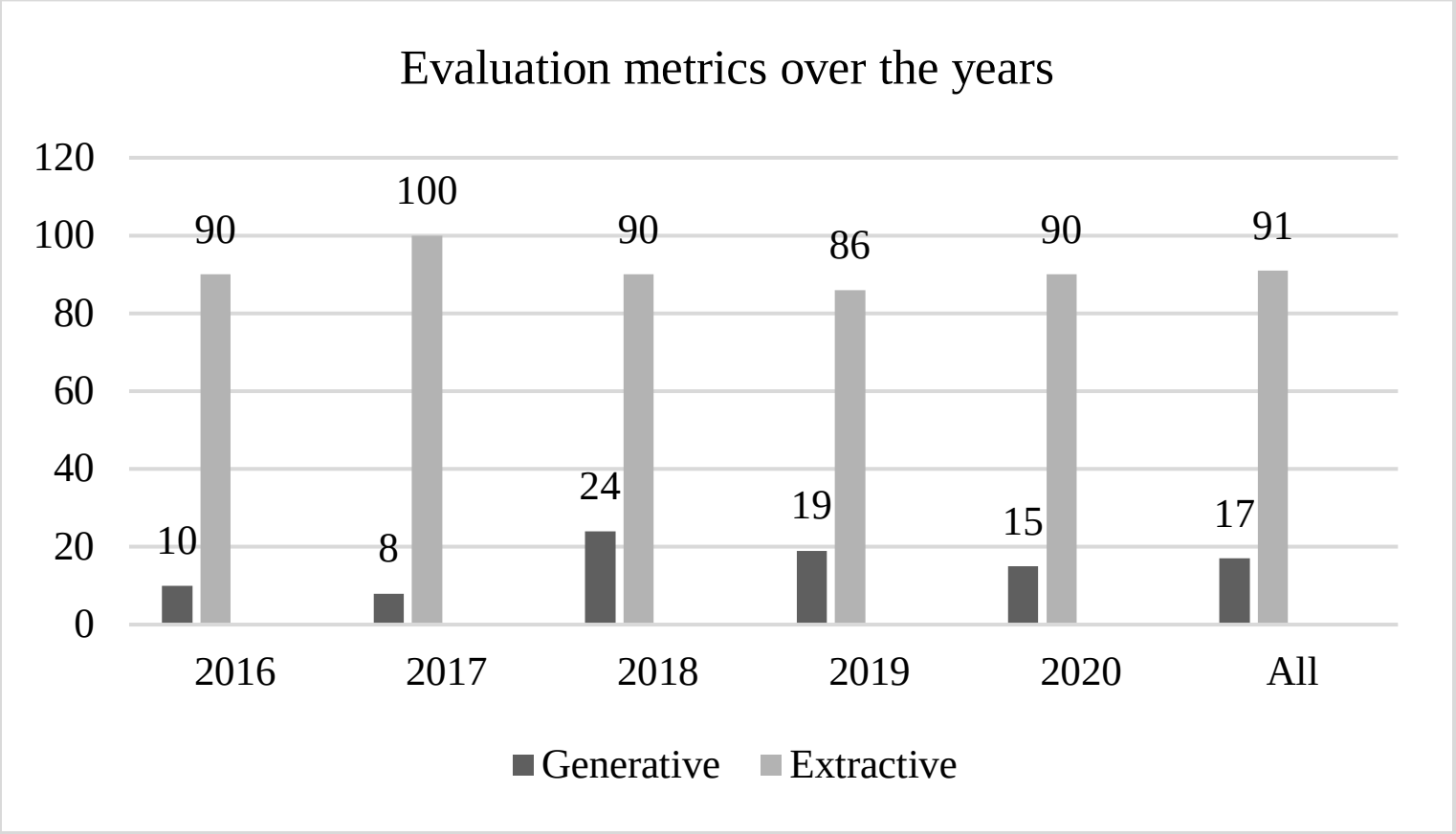

**Figure 3.** Ratio of reviewed papers (%) for extractive/generative evaluation metrics

## 7. Research contribution

The contribution of MRC research can be grouped into four categories: developing new model structures, creating new datasets, combining with other tasks and improvement, and introducing new evaluation measures. Table 9 shows the statistics of these categories. Note that some studies have more than one contribution type, so the sum of some rows is greater than 100%. For example, Ma, Jurczyk, and Choi (2018) introduced a new dataset from the "Friends" sitcom transcripts and developed a new model architecture as well. For more details, refer to Table A6.

### *7.1 Developing new model structures*

Many MRC papers have focused on developing new model structures to address the weaknesses of previous models. Most of them developed new internal structures (Seo et al. 2017; Cui, Chen, et al. 2017; Dhingra et al. 2017; Kobayashi et al. 2016; Xiong, Zhong, and Socher 2017; Cui et al. 2016; Kadlec et al. 2016; Hu, Peng, Huang, et al. 2018; Shen et al. 2017; Huang et al. 2018). Some others changed the system inputs. For example, in Chen et al. 2017) study, in addition to word embedding, linguistic features such as NER and POS embeddings have also been used as the input to the model. Also, some papers introduced a new way of entering the input into the system. For example, Hewlett, Jones, and Lacoste (2017) proposed breaking the context into overlapping windows and entering each window as an input to the system.

### *7.2 Creating new datasets*

One of the main reasons for advancing the MRC research in recent years is the introduction of rich datasets. Many studies have focused on creating new datasets with new features in recent years (Nguyen et al. 2016; Trischler et al. 2017; Rajpurkar et al. 2016; Šuster and Daelemans 2018; Joshi et al. 2017; Lai et al. 2017; He et al. 2018). The main trend is to develop multi-document datasets, abstractive style outputs, and more complex questions that require more advanced reasoning. Also, some papers focus on customizing the available datasets instead of creating new ones. For example, Horbach et al. (2020) proposed to turn existing datasets, such as SQuAD and NewsQA, into interactive datasets. Some other papers added annotations to the existing datasets to provide interpretable clues for investigating the models' behavior as well as to prevent models from exploiting biases and annotation artifacts. For example, SQuAD 2.0 (Rajpurkar, Jia, and Liang 2018) was created by adding unanswerable questions to SQuAD, and SQuAD2-CR (Lee, Hwang, and Cho 2020) was developed by adding causal and relational annotations to unanswerable questions in SQuAD 2.0. Similarly, $R^4C$ (Inoue, Stenetorp, and Inui 2020) was created by adding derivations to questions in hotpotQA.

### *7.3 Combining with other tasks*

Simultaneous learning of multiple tasks (multi-task learning) (Collobert and Weston 2008) and exploiting the learned knowledge from one task in another task (transfer learning) (Ruder 2019) have been promising directions for obtaining better results, especially in the data-poor setting. As an example, Wang, Yuan, and Trischler (2017) trained their MRC task with a question generation task and achieved better results. Besides these approaches, some papers exploit other task solutions as sub-modules in their MRC system. As an example, Yin, Ebert, and Schütze (2016) used a question classifier and a natural language inference (NLI) system as two sub-modules in their MRC system.

### *7.4 Introducing new evaluation measures*

Reliable assessment of an MRC system is still a challenging topic. While some systems go beyond human performance in specific datasets such as SQuAD by the current measures (Rajpurkar et al. 2016), further investigations show that these systems fail to achieve a thorough and true understanding of human language (Jia and Liang 2017; Wang and Bansal 2018). In these papers, the passage is successfully edited to mislead the model. These papers can be seen as a measure to evaluate the true comprehension of systems. Also, some papers have evaluated the required comprehension and reasoning capabilities for solving the MRC problem in available datasets (Chen, Bolton, and Manning 2016; Sugawara et al. 2018).

**Table 9.** Statistics of different research contributions to MRC task in the reviewed papers.

| YEAR | MODEL STRUCTURE | DATASET | OTHER TASKS | EVALUATION MEASURE |
|---|---|---|---|---|
| 2016 | 50% | 50% | 21% | 7% |
| 2017 | 54% | 14% | 23% | 6% |
| 2018 | 71% | 31% | 14% | 5% |
| 2019 | 68% | 20% | 24% | 11% |
| 2020 | 57% | 20% | 29% | 31% |
| All (249) | 61% | 23% | 21% | 12% |

## 8. Hot MRC papers

Table 10 shows the top papers in different years from 2016 to 2020 based on the number of citations in the Google Scholar service until September, 2020. For all years but 2020, ten papers have been selected; while for the year 2020, just five papers have been chosen because the number of citations in this year is not enough in the time of writing the paper. According to this table, hot papers often introduce a new successful model structure or a new dataset.

**Table 10.** Hot papers based on the number of citations in the Google Scholar service until September, 2020.

| TITLE | PUBLICATION VENUE | YEAR | CONTRIBUTION | CITATIONS |
|---|---|---|---|---|
| Squad: 100,000+ questions for machine comprehension of text (Rajpurkar et al. 2016) | EMLP | 2016 | Dataset | 2068 |
| The goldilocks principle: Reading children's books with explicit memory representations (Hill et al. 2016) | ICLR | 2016 | Dataset | 426 |
| A thorough examination of the CNN/Daily Mail reading comprehension task (Chen, Bolton, and Manning 2016) | ACL | 2016 | Evaluation dataset | 420 |
| MS MARCO: A human generated machine reading comprehension dataset (Nguyen et al. 2016) | CoCo@NIPS | 2016 | Dataset | 407 |
| Text understanding with the attention sum reader network (Kadlec et al. 2016) | ACL | 2016 | Model Structure (AS) | 241 |
| Who did what: A large-scale person-centered cloze dataset (Onishi et al. 2016) | EMNLP | 2016 | Dataset | 97 |
| Attention-based convolutional neural network for machine comprehension (Yin, Ebert, and Schütze 2016) | NAACL | 2016 | Model Structure (HABCNN), Knowledge Transfer | 78 |
| Consensus attention-based neural networks for Chinese reading comprehension (Cui et al. 2016) | COLING | 2016 | Model structure (CAS Reader), Dataset | 58 |
| Dynamic entity representation with max-pooling improves machine reading (Kobayashi et al. 2016) | ACL | 2016 | Model Structure (DER Network) | 40 |
| Employing external rich Knowledge for machine comprehension (Wang et al. 2016) | IJCAI | 2016 | Model Structure, Knowledge Transfer | 12 |
| Bidirectional attention flow for machine comprehension (Seo et al. 2017) | ICLR | 2017 | Model structure (BiDAF) | 1091 |
| Reading wikipedia to answer open-domain questions (Chen et al. 2017) | ACL | 2017 | Model structure (DrQA) | 681 |
| Adversarial examples for evaluating reading comprehension systems (Jia and Liang 2017) | EMNLP | 2017 | Evaluation measure | 547 |
| Dynamic coattention networks for question answering (Xiong, Zhong, and Socher 2017) | ICLR | 2017 | Model structure (DCN) | 448 |
| Gated self-matching networks for reading comprehension and question answering (Wang et al. 2017) | ACL | 2017 | Model structure (R-NET) | 430 |
| Triviaqa: a large scale distantly supervised challenge dataset for reading comprehension (Joshi et al. 2017) | ACL | 2017 | Dataset | 418 |
| Machine comprehension using match-LSTM and answer pointer (Wang and Jiang 2017) | ICLR | 2017 | Model structure (Match-LSTM and Answer Pointer) | 392 |
| NewsQA: A machine comprehension dataset (Trischler et al. 2017) | RepL4NLP | 2017 | Dataset | 302 |
| Attention-over-attention neural networks for reading comprehension (Cui, Chen, et al. 2017) | ACL | 2017 | Model structure (AoA Reader) | 291 |
| Gated-attention readers for text comprehension (Dhingra et al. 2017) | ACL | 2017 | Model structure (GA-Reader) | 288 |
| Know what you don't know: unanswerable questions for squad (Rajpurkar, Jia, and Liang 2018) | ACL | 2018 | Dataset | 529 |
| QANet: combining local convolution with global self-attention for reading comprehension (Yu et al. 2018) | ICLR | 2018 | Model structure (QANet) | 407 |
| HotpotQA: A dataset for diverse, explainable multi-hop question answering (Yang et al. 2018) | EMNLP | 2018 | Dataset | 260 |
| Simple and effective multi-paragraph reading comprehension (Clark and Gardner 2018) | ACL | 2018 | Model structure | 212 |
| Reinforced mnemonic reader for machine reading comprehension (Hu, Peng, Huang, et al. 2018) | IJCAI | 2018 | Model structure (R.M-Reader) | 184 |
| QuAC: question answering in context (Choi et al. 2018) | EMNLP | 2018 | Dataset | 173 |
| The narrativeqa reading comprehension challenge (Kočiský et al. 2018) | ACL | 2018 | Dataset | 172 |
| Constructing datasets for multi-hop reading comprehension across documents (Welbl, Stenetorp, and Riedel 2018) | ACL | 2018 | Dataset | 170 |
| R3: reinforced ranker-reader for open-domain question answering (Wang, Yu, Guo, et al. 2018) | AAAI | 2018 | Model structure (R3) | 150 |
| Stochastic answer networks for machine reading comprehension (Liu, Shen, et al. 2018) | ACL | 2018 | Model structure (SAN) | 130 |
| Natural questions: a benchmark for question answering research (Kwiatkowski et al. 2019) | ACL | 2019 | Dataset | 192 |
| DROP: A reading comprehension benchmark requiring discrete reasoning over paragraphs (Dua et al. 2019) | NAACL-HLT | 2019 | Model Structure (NAQANET), Dataset | 108 |
| Read + verify: Machine reading comprehension with unanswerable questions (Hu, Wei, et al. 2019) | AAAI | 2019 | Model Structure (RMR) | 69 |
| Improving machine reading comprehension with general reading strategies (Sun et al. 2019) | NAACL-HLT | 2019 | Model Structure, Knowledge Transfer | 54 |
| MultiQA: An empirical investigation of generalization and transfer in reading comprehension (Talmor and Berant 2019) | ACL | 2019 | Evaluation Measure | 49 |
| Multi-step retriever-reader interaction for scalable open-domain question answering (Das et al. 2019) | ICLR | 2019 | Model Structure | 47 |
| Cognitive graph for multi-hop reading comprehension at scale (Ding et al. 2019) | ACL | 2019 | Model Structure (CogQA) | 43 |
| Cosmos QA: Machine reading comprehension with contextual commonsense ceasoning (Huang, Le Bras, et al. 2019) | EMNLP-IJCNLP | 2019 | Dataset, Model Structure | 32 |
| Multi-hop reading comprehension through question decomposition and rescoring (Min et al. 2019) | ACL | 2019 | Model Structure | 28 |
| MRQA 2019 shared task: Evaluating generalization in reading comprehension (Fisch et al. 2019) | ACL | 2019 | Evaluation Measure | 27 |
| SG-Net: syntax-guided machine reading comprehension (Zhang, Wu, et al. 2020) | AAAI | 2020 | Model Structure (SG-Net), Knowledge Transfer | 17 |
| Select, answer and explain: Interpretable multi-Hop reading comprehension over multiple documents (Tu et al. 2020) | AAAI | 2020 | Model Structure (SAE), Knowledge Transfer | 10 |
| DCMN+: Dual co-matching network for multi-choice reading comprehension (Zhang, Zhao, et al. 2020) | AAAI | 2020 | Model Structure (DCMN+) | 9 |
| Assessing the benchmarking capacity of machine reading comprehension datasets (Sugawara et al. 2020) | AAAI | 2020 | Evaluation Measure | 8 |
| Mmm: Multi-stage multi-task learning for multi-choice reading comprehension (Jin et al. 2020) | AAAI | 2020 | Model Structure (MMM), Knowledge Transfer | 5 |

## 9. Future trends and opportunities

The MRC task has witnessed a great progress in recent years. Especially, like other NLP tasks, fine-tuning the pre-trained language models like BERT (Devlin et al. 2018) and XLNet (Yang, Dai, et al. 2019) on the target task has achieved an impressive success in MRC task, such that many state-of-the-art systems use these language models. However, they suffer from some shortcomings, which make them far from the real reading comprehension. In the following, we list some of these challenges and new trends in the MRC field:

- *Out-of-domain distributions*: Despite the high accuracy of the current MRC models on test samples from their training distribution, they are too fragile for out-of-domain distributed data. To address this shortcoming, some of recent papers focused on improving the generalization capability of MRC models (Nishida et al. 2020; Wang, Gan, et al. 2019; Fisch et al. 2019).
- *Multi-document MRC:* One important challenge in the MRC task is the multi-hop reasoning, which is to infer the answer from multiple texts. These texts can be either several paragraphs of a document (Frermann 2019; Tay et al. 2019) or heterogeneous paragraphs from multiple documents (Tu et al. 2020). One of the new trends is to use the graph structures, such as graph neural networks, for multi-hop reasoning (Tu et al. 2020; Tu et al. 2019; Song et al. 2020; Ding et al. 2019).
- *Numerical reasoning:* Many questions in real-world applications require numerical inference including addition, subtraction, comparison, and so on. For example, consider the following question from DROP dataset (Dua et al. 2019) which needs a subtraction: "How many more dollars was the Untitled (1981) painting sold for than the 12 million dollar estimation?" Developing MRC models capable of numerical reasoning has become more popular in recent years, especially after the creation of numerical datasets such as DROP (Dua et al. 2019; Ran et al. 2019; Chen et al. 2020).
- *No-answer questions:* One of the new trends that makes the MRC systems more usable in real-world applications is to enable models to identify the questions which cannot be answered using the given context. With the development of datasets containing this kind of questions, such as SQuAD 2.0 (Rajpurkar, Jia, and Liang 2018) and Natural Questions (Kwiatkowski et al. 2019), more attention has been paid to this issue (Back et al. 2020; Liu, Gong, et al. 2020; Zhang, Wu, et al. 2020).
- *Non-factoid questions:* Answering non-factoid questions, such as *why* and *opinion* questions, often requires generating answers rather than selecting a span of context. The accuracy of the existing models in answering these questions is still far from the desired level. In recent years, several datasets which contain non-factoid questions have attracted more attentions to this kind of questions (Berzak, Malmaud, and Levy 2020; Gupta et al. 2019; Saha et al. 2018).
- *Low-resource language datasets and models:* It is worth noting that most available datasets are in resource-rich languages, such as English and Chinese. Creating new datasets and models for low-resource languages and developing them in the multi-lingual or multi-task setting can also be seen as a new trend in this field (Amirkhani et al. 2020; Yuan, Shou, et al. 2020; Gupta and Khade 2020; Jing, Xiong, and Zhen 2019).

## 10. Conclusion

Machine reading comprehension, as a hot research topic in NLP, focuses on reading the document(s) and answering questions about it. The ideal goal of an MRC system is to gain a comprehensive understanding of text documents to be able to reason and answer related questions. In this paper, we presented an overview of different aspects of recent MRC studies, including approaches, internal architecture, input/output type, research contributions, and evaluation measures. We reviewed 241 papers from 2016 to 2020 to investigate recent studies and find new trends.

Based on the question type, MRC papers are categorized to factoid, non-factoid, and yes/no questions. The input context is also categorized to single or multiple passages. According to statistics, a trend toward non-factoid questions and multiple passages is obvious in recent years. The output types are categorized to extractive and abstractive outputs. From another point of view, the output types are classified as quiz, cloze, and detail styles. The statistics show that even though the extractive outputs have been more popular, the abstractive outputs are becoming more popular in recent years.

We also reviewed the developed datasets along with their features, including data volume, domain, question type, answer type, context type, collection method, and data language. The number of developed datasets has increased in recent years, and they are in general more challenging than previous datasets. Regarding research contributions, some papers develop new model structures, some introduce new datasets, some combine MRC task with other tasks, and others introduce new evaluation measures. Among these, the majority of papers develop new model structures or introduce new datasets. We also presented the most-cited papers, which show the most popular datasets and models in the MRC literature. Finally, we mentioned the future trends and important challenges of available models including the issues related to out-of-domain distributions, multi-document MRC, numerical reasoning, no-answer questions, non-factoid questions, and low-resource languages.

## Appendix

**Table A1.** Reviewed papers categorized based on their input/output

| | | | |
|---|---|---|---|
| INPUT | QUESTION | FACTOID | (Chen et al. 2017), (Yang, Hu, et al. 2017), (Clark and Gardner 2018), (Weissenborn, Wiese, and Seiffe 2017), (Choi, Hewlett, Uszkoreit, et al. 2017), (Hu, Peng, Huang, et al. 2018), (Shen et al. 2017), (Wang et al. 2017), (Liu, Shen, et al. 2018), (Huang et al. 2018), (Tan, Wei, Yang, et al. 2018), (Zhang et al. 2017), (Liu et al. 2017), (Gong and Bowman 2018), (Wang, Yuan, and Trischler 2017), (Wang, Yu, Guo, et al. 2018), (Hewlett, Jones, and Lacoste 2017), (Golub et al. 2017), (Swayamdipta, Parikh, and Kwiatkowski 2018), (Xiong, Zhong, and Socher 2018), (Xie and Xing 2017), (Salant and Berant 2018), (Wang, Yu, Jiang, et al. 2018), (Mihaylov, Kozareva, and Frank 2017), (Greco et al. 2016), (Cui, Chen, et al. 2017), (Dhingra et al. 2017), (Wang, Yu, Chang, et al. 2018), (Kundu and Ng 2018b), (Prakash, Tripathy, and Banu 2018), (Lin, Liu, and Li 2018), (Liu, Wei, et al. 2018), (Kundu and Ng 2018a), (Hu, Peng, Wei, et al. 2018), (Bauer, Wang, and Bansal 2018), (Hoang, Wiseman, and Rush 2018), (Aghaebrahimian 2018), (Seo et al. 2018), (Back et al. 2018), (Li, Li, and Lv 2018), (Zhang et al. 2018), (Liu, Huang, et al. 2018), (Wang, Yan, and Wu 2018), (Wang, Liu, Liu, et al. 2018), (Liu, Zhao, et al. 2018), (Dhingra, Pruthi, and Rajagopal 2018), (Yang, Dhingra, et al. 2017), (Seo et al. 2017), (Wang and Jiang 2017), (Chen, Bolton, and Manning 2016), (Kobayashi et al. 2016), (Trischler et al. 2017), (Xiong, Zhong, and Socher 2017), (Cui et al. 2016), (Yin, Ebert, and Schütze 2016), (Wang et al. 2016), (Kadlec, Bajgar, and Kleindienst 2016), (Kadlec et al. 2016), (Liu and Perez 2017), , (Munkhdalai and Yu 2017), (Ma, Jurczyk, and Choi 2018), (Chaturvedi, Pandit, and Garain 2018), (Indurthi et al. 2018), (Ghaeini et al. 2018), (Sheng, Lan, and Wu 2018), (Min et al. 2018), (Sugawara et al. 2018), (Tay, Luu, and Hui 2018), (Gupta et al. 2018), (Ke et al. 2018), (Du and Cardie 2018), (Tan, Wei, Zhou, et al. 2018), (Wang, Liu, Xiao, et al. 2018), (Mihaylov and Frank 2018), (Yu et al. 2018), (Song et al. 2018), (Nishida et al. 2018), (Wang and Bansal 2018), (Sachan and Xing 2018), (Htut, Bowman, and Cho 2018), (Miao, Liu, and Gao 2019a), (Yan et al. 2019), (Hu, Peng, et al. 2019a), (Wang, Gan, et al. 2019), (Jin, Yang, and Zhu 2019), (Angelidis et al. 2019), (Ding et al. 2019), (Chen, Cui, et al. 2019), (Takahashi et al. 2019), (Huang, Le Bras, et al. 2019), (Cui, Che, et al. 2019), (Mihaylov and Frank 2019), (Li, Zhang, Liu, et al. 2019), (Wang, Yu, Guo, et al. 2019), (Dua et al. 2019), (Zhang, Zhao, et al. 2019), (Xu, Liu, Chen, et al. 2019), (Yang, Wang, et al. 2019), (Wang, Yu, Sun, et al. 2019), (Wang and Jiang 2019), (Jiang et al. 2019), (Pang et al. 2019),(Su et al. 2019), (Andor et al. 2019), (Tang et al. 2019), (Sharma and Roychowdhury 2019), (Sun et al. 2019), (Wu et al. 2019), (Bi et al. 2019), (Xia, Wu, and Yan 2019), (Yu, Zha, and Yin 2019), (Dehghani et al. 2019), (Min et al. 2019), (Tu et al. 2019), (Tang, Cai, and Zhuo 2019), (Yuan et al. 2019), (Ren et al. 2019), (Nishida, Saito, et al. 2019), (Xu, Liu, Shen, et al. 2019), (Ran et al. 2019), (Miao, Liu, and Gao 2019b), (Li, Zhang, Zhu, et al. 2019), (Osama, El-Makky, and Torki 2019), (Liu, Zhang, Zhang, and Wang 2019), (Hu, Wei, et al. 2019), (Hu, Peng, et al. 2019b), (Huang, Tang, et al. 2019), (Nie, Wang, and Bansal 2019), (Tay et al. 2019), (Xiong et al. 2019), (Fremann 2019), (Zhuang and Wang 2019), (Li, Li, and Liu 2019), (Wang, Yao, et al. 2019), (Park, Lee, and Song 2019), (Nishida, Nishida, et al. 2019), (Lee, Kim, and Park 2019), (Das et al. 2019), (Guo et al. 2020), (Wang, Zhang, et al. 2020), (Nakatsuji and Okui 2020), (Zhou, Luo, and Wu 2020), (Zhang, Zhao, et al. 2020), (Lee and Kim 2020), (Yang, Kang, and Seo 2020), (Wu and Xu 2020), (Zhang and Wang 2020), (Zhang, Luo, et al. 2020), (Liu, Gong, et al. 2020), (Tu et al. 2020), (Jin et al. 2020), (Chen et al. 2020), (Niu et al. 2020), (Pappas et al. 2020), (Zheng et al. 2020), (Song et al. 2020), (Ren, Cheng, and Su 2020), (Back et al. 2020), (Zhang, Wu, et al. 2020) |
| | | NON-FACTOID | (Min, Seo, and Hajishirzi 2017), (Choi, Hewlett, Uszkoreit, et al. 2017), (Tan, Wei, Yang, et al. 2018), , (Hewlett, Jones, and Lacoste 2017), (Wang, Yu, Jiang, et al. 2018), (Wang, Yu, Chang, et al. 2018),(Lin, Liu, and Li 2018), (Liu, Wei, et al. 2018), (Bauer, Wang, and Bansal 2018), (Aghaebrahimian 2018), (Li, Li, and Lv 2018), (Zhang et al. 2018), (Wang, Liu, Liu, et al. 2018), (Liu, Zhao, et al. 2018),(Dhingra, Pruthi, and Rajagopal 2018), (Wang et al. 2016), (Tay, Luu, and Hui 2018), (Miao, Liu, and Gao 2019a), (Yan et al. 2019), (Jin, Yang, and Zhu 2019), (Chen, Cui, et al. 2019), (Takahashi et al. 2019), (Huang, Le Bras, et al. 2019), (Cui, Che, et al. 2019), (Mihaylov and Frank 2019), (Li, Zhang, Liu, et al. 2019), (Wang, Yu, Guo, et al. 2019), (Wang, Yu, Sun, et al. 2019), (Su et al. 2019), (Sharma and Roychowdhury 2019), (Sun et al. 2019), (Xia, Wu, and Yan 2019), (Yu, Zha, and Yin 2019), (Tang, Cai, and Zhuo 2019), (Ren et al. 2019), (Nishida, Saito, et al. 2019), (Miao, Liu, and Gao 2019b), (Li, Zhang, Zhu, et al. 2019), (Osama, El-Makky, and Torki 2019) , (Tay et al. 2019), (Fremann 2019), (Lee, Kim, and Park 2019), (Zhou, Luo, and Wu 2020), (Zhang, Zhao, et al. 2020), (Liu, Gong, et al. 2020), (Jin et al. 2020), (Zheng et al. 2020), (Zhang, Luo, et al. 2020), (Zhang, Wu, et al. 2020) |
| | | YES/NO | (Liu, Wei, et al. 2018), (Li, Li, and Lv 2018), (Wang, Liu, Liu, et al. 2018), (Zhang et al. 2018), (Liu, Zhao, et al. 2018), (Hu, Peng, et al. 2019a), (Nakatsuji and Okui 2020), (Niu et al. 2020), (Liu, Gong, et al. 2020), (Tu et al. 2020), (Zheng et al. 2020) |
| | CONTEXT | SINGLE PARAGRAPH | (Chen et al. 2017),(Yang, Hu, et al. 2017), (Min, Seo, and Hajishirzi 2017), (Weissenborn, Wiese, and Seiffe 2017), (Choi, Hewlett, Uszkoreit, et al. 2017), (Hu, Peng, Huang, et al. 2018), (Shen et al. 2017), (Wang et al. 2017), (Liu, Shen, et al. 2018), (Huang et al. 2018), (Zhang et al. 2017), (Liu et al. 2017), (Gong and Bowman 2018), (Wang, Yuan, and Trischler 2017), (Hewlett, Jones, and Lacoste 2017), (Golub et al. 2017), (Xiong, Zhong, and Socher 2018), (Xie and Xing 2017), (Salant and Berant 2018), (Wang, Yu, Jiang, et al. 2018), (Mihaylov, Kozareva, and Frank 2017), (Cui, Chen, et al. 2017), (Dhingra et al. 2017) ,(Wang, Yu, Chang, et al. 2018), (Kundu and Ng 2018b), (Prakash, Tripathy, and Banu 2018), (Lin, Liu, and Li 2018), (Kundu and Ng 2018a), (Hu, Peng, Wei, et al. 2018),(Bauer, Wang, and Bansal 2018), (Hoang, Wiseman, and Rush 2018), (Aghaebrahimian 2018), (Back et al. 2018), (Seo et al. 2018), (Wang, Yan, and Wu 2018), (Liu, Zhao, et al. 2018), (Dhingra, Pruthi, and Rajagopal 2018), (Yang, Dhingra, et al. 2017), (Seo et al. 2017), (Wang and Jiang 2017), |
| | | | (Chen, Bolton, and Manning 2016), (Kobayashi et al. 2016), (Trischler et al. 2017), (Xiong, Zhong, and Socher 2017), (Cui et al. 2016), (Yin, Ebert, and Schütze 2016), (Wang et al. 2016), (Kadlec, Bajgar, and Kleindienst 2016), (Kadlec et al. 2016), (Liu and Perez 2017), , (Munkhdalai and Yu 2017), (Ma, Jurczyk, and Choi 2018), (Chaturvedi, Pandit, and Garain 2018), (Indurthi et al. 2018), (Ghaeini et al. 2018), (Sheng, Lan, and Wu 2018), (Min et al. 2018), (Sugawara et al. 2018), (Tay, Luu, and Hui 2018), (Gupta et al. 2018), (Du and Cardie 2018), (Tan, Wei, Zhou, et al. 2018), (Mihaylov and Frank 2018), (Yu et al. 2018), (Song et al. 2018), (Nishida et al. 2018), (Wang and Bansal 2018), (Sachan and Xing 2018), (Hu, Peng, et al. 2019a), (Wang, Gan, et al. 2019), (Jin, Yang, and Zhu 2019), (Chen, Cui, et al. 2019), (Takahashi et al. 2019), (Huang, Le Bras, et al. 2019), (Cui, Che, et al. 2019), (Mihaylov and Frank 2019), (Li, Zhang, Liu, et al. 2019), (Dua et al. 2019), (Zhang, Zhao, et al. 2019), (Xu, Liu, Chen, et al. 2019), (Yang, Wang, et al. 2019), (Wang and Jiang 2019), (Su et al. 2019), (Andor et al. 2019), (Sharma and Roychowdhury 2019) , (Wu et al. 2019), (Xia, Wu, and Yan 2019), (Xu, Liu, Shen, et al. 2019), (Ran et al. 2019), (Li, Zhang, Zhu, et al. 2019), (Osama, El-Makky, and Torki 2019), (Liu, Zhang, Zhang, and Wang 2019), (Huang, Tang, et al. 2019), (Hu, Wei, et al. 2019), (Li, Li, and Liu 2019), (Park, Lee, and Song 2019), (Lee, Kim, and Park 2019), (Guo et al. 2020), (Wang, Zhang, et al. 2020), (Zhou, Luo, and Wu 2020), (Zhang, Zhao, et al. 2020), (Lee and Kim 2020), (Wu and Xu 2020), (Zhang and Wang 2020), (Zhang, Luo, et al. 2020), (Jin et al. 2020), (Chen et al. 2020), (Niu et al. 2020), (Pappas et al. 2020), (Back et al. 2020), (Zhang, Wu, et al. 2020) |
| | | Multi- paragraph | (Clark and Gardner 2018), (Tan, Wei, Yang, et al. 2018), (Wang, Yu, Guo, et al. 2018), (Swayamdipta, Parikh, and Kwiatkowski 2018), (Wang, Yu, Jiang, et al. 2018),(Greco et al. 2016), (Liu, Wei, et al. 2018), (Aghaebrahimian 2018), (Li, Li, and Lv 2018), (Zhang et al. 2018), (Liu, Huang, et al. 2018), (Wang, Liu, Liu, et al. 2018), (Nguyen et al. 2016), (Tay, Luu, and Hui 2018), (Ke et al. 2018), (Wang, Liu, Xiao, et al. 2018), (Yu et al. 2018), (Sachan and Xing 2018), (Htut, Bowman, and Cho 2018), (Miao, Liu, and Gao 2019a), (Yan et al. 2019), (Wang, Gan, et al. 2019), (Angelidis et al. 2019), (Ding et al. 2019), (Wang, Yu, Guo, et al. 2019), (Wang, Yu, Sun, et al. 2019), (Jiang et al. 2019), (Pang et al. 2019), (Tang et al. 2019), (Sun et al. 2019), (Bi et al. 2019), (Yu, Zha, and Yin 2019), (Dehghani et al. 2019), (Min et al. 2019), (Tu et al. 2019), (Tang, Cai, and Zhuo 2019), (Yuan et al. 2019), (Ren et al. 2019), (Nishida, Saito, et al. 2019), (Miao, Liu, and Gao 2019b), (Hu, Peng, et al. 2019b), (Nie, Wang, and Bansal 2019), (Tay et al. 2019), (Xiong et al. 2019), (Frermann 2019), (Zhuang and Wang 2019), (Wang, Yao, et al. 2019), (Nishida, Nishida, et al. 2019), (Das et al. 2019), (Nakatsuji and Okui 2020), (Yang, Kang, and Seo 2020), (Liu, Gong, et al. 2020), (Tu et al. 2020), (Ren, Cheng, and Su 2020), (Zheng et al. 2020), (Song et al. 2020) |
| Output | - | Extractive | (Yang, Hu, et al. 2017), (Yang, Hu, et al. 2017), (Clark and Gardner 2018), (Min, Seo, and Hajishirzi 2017), (Weissenborn, Wiese, and Seiffe 2017), (Hu, Peng, Huang, et al. 2018),(Shen et al. 2017), (Wang et al. 2017), (Liu, Shen, et al. 2018), (Huang et al. 2018), (Zhang et al. 2017), (Liu et al. 2017), (Gong and Bowman 2018), (Wang, Yuan, and Trischler 2017), (Wang, Yu, Guo, et al. 2018), (Golub et al. 2017), (Swayamdipta, Parikh, and Kwiatkowski 2018), (Xiong, Zhong, and Socher 2018), (Xie and Xing 2017), (Salant and Berant 2018), (Wang, Yu, Jiang, et al. 2018), (Mihaylov, Kozareva, and Frank 2017), (Cui, Chen, et al. 2017), (Dhingra et al. 2017), (Kundu and Ng 2018b), (Prakash, Tripathy, and Banu 2018), (Liu, Wei, et al. 2018), (Kundu and Ng 2018a), (Hu, Peng, Wei, et al. 2018), (Hoang, Wiseman, and Rush 2018), (Aghaebrahimian 2018), (Back et al. 2018), (Seo et al. 2018), (Li, Li, and Lv 2018), (Liu, Huang, et al. 2018), (Wang, Yan, and Wu 2018), (Wang, Liu, Liu, et al. 2018), (Dhingra, Pruthi, and Rajagopal 2018), (Yang, Dhingra, et al. 2017), (Seo et al. 2017), (Wang and Jiang 2017), (Chen, Bolton, and Manning 2016), (Kobayashi et al. 2016), (Trischler et al. 2017), (Xiong, Zhong, and Socher 2017), (Cui et al. 2016), (Yin, Ebert, and Schütze 2016), (Wang et al. 2016), (Kadlec, Bajgar, and Kleindienst 2016), (Kadlec et al. 2016), (Liu and Perez 2017), (Munkhdalai and Yu 2017), (Ma, Jurczyk, and Choi 2018), (Chaturvedi, Pandit, and Garain 2018), (Ghaeini et al. 2018), (Sheng, Lan, and Wu 2018), (Min et al. 2018), (Sugawara et al. 2018), (Tay, Luu, and Hui 2018), (Gupta et al. 2018), (Ke et al. 2018), (Du and Cardie 2018), (Tan, Wei, Zhou, et al. 2018), (Wang, Liu, Xiao, et al. 2018), (Mihaylov and Frank 2018), (Yu et al. 2018), (Song et al. 2018), (Nishida et al. 2018), (Wang and Bansal 2018), (Sachan and Xing 2018), (Htut, Bowman, and Cho 2018),(Yan et al. 2019), (Hu, Peng, et al. 2019a), (Wang, Gan, et al. 2019), (Jin, Yang, and Zhu 2019), (Angelidis et al. 2019), (Ding et al. 2019), (Takahashi et al. 2019), (Cui, Che, et al. 2019), (Mihaylov and Frank 2019), (Li, Zhang, Liu, et al. 2019), (Wang, Yu, Guo, et al. 2019), (Dua et al. 2019), (Zhang, Zhao, et al. 2019), (Xu, Liu, Chen, et al. 2019), (Yang, Wang, et al. 2019), (Wang and Jiang 2019), (Jiang et al. 2019), (Su et al. 2019), (Andor et al. 2019), (Pang et al. 2019), (Tang et al. 2019), (Wu et al. 2019), (Dehghani et al. 2019), (Min et al. 2019), (Tu et al. 2019), (Yuan et al. 2019), (Ren et al. 2019), (Xu, Liu, Shen, et al. 2019), (Ran et al. 2019), (Li, Zhang, Zhu, et al. 2019), (Osama, El-Makky, and Torki 2019), (Liu, Zhang, Zhang, and Wang 2019), (Huang, Tang, et al. 2019), (Hu, Peng, et al. 2019b), (Nie, Wang, and Bansal 2019), (Xiong et al. 2019), (Frermann 2019), (Hu, Wei, et al. 2019), (Zhuang and Wang 2019), (Li, Li, and Liu 2019), (Park, Lee, and Song 2019), (Nishida, Nishida, et al. 2019), (Lee, Kim, and Park 2019), (Das et al. 2019), (Guo et al. 2020), (Wang, Zhang, et al. 2020), (Zhang, Zhao, et al. 2020), (Lee and Kim 2020), (Yang, Kang, and Seo 2020), (Wu and Xu 2020), (Zhang and Wang 2020), (Liu, Gong, et al. 2020), (Tu et al. 2020), (Ren, Cheng, and Su 2020), (Niu et al. 2020), (Pappas et al. 2020), (Zheng et al. 2020), (Song et al. 2020), (Chen et al. 2020), (Back et al. 2020), (Zhang, Wu, et al. 2020) |
| | | Abstractive | (Choi, Hewlett, Uszkoreit, et al. 2017), (Tan, Wei, Yang, et al. 2018), (Hewlett, Jones, and Lacoste 2017), (Greco et al. 2016),(Wang, Yu, Chang, et al. 2018),(Lin, Liu, and Li 2018), (Bauer, Wang, and Bansal 2018), (Zhang et al. 2018), (Liu, Zhao, et al. 2018), , (Indurthi et al. 2018), (Ke et al. 2018), (Miao, Liu, and Gao 2019a), (Hu, Peng, et al. 2019a), (Wang, Gan, et al. 2019), (Chen, Cui, et al. 2019), (Huang, Le Bras, et al. 2019), (Dua et al. 2019), (Wang, Yu, Sun, et al. 2019), (Andor et al. 2019), (Sharma and Roychowdhury 2019), (Sun et al. 2019), (Bi et al. 2019)¸ (Xia, Wu, and Yan 2019), (Yu, Zha, and Yin 2019), (Tang, Cai, and Zhuo 2019), (Nishida, Saito, et al. 2019), (Ran et al. 2019), (Miao, Liu, and Gao 2019b), (Li, Zhang, Zhu, et al. 2019), (Liu, Zhang, Zhang, and Wang 2019), (Tay et al. 2019), (Wang, Yao, et al. 2019), (Niu et al. 2020), (Nakatsuji and Okui 2020), (Zhou, Luo, and Wu 2020), (Zhang, Zhao, et al. 2020), (Zhang, Luo, et al. |
| | | | 2020), (Liu, Gong, et al. 2020), (Tu et al. 2020), (Jin et al. 2020), (Zheng et al. 2020), (Chen et al. 2020), (Zhang, Wu, et al. 2020) |
| | | QUIZ | (Wang, Yu, Jiang, et al. 2018), (Greco et al. 2016), (Wang, Yu, Chang, et al. 2018), (Lin, Liu, and Li 2018), (Zhang et al. 2018), (Liu, Zhao, et al. 2018), (Yang, Dhingra, et al. 2017), (Yin, Ebert, and Schütze 2016), (Wang et al. 2016), (Chaturvedi, Pandit, and Garain 2018), (Sheng, Lan, and Wu 2018), (Sugawara et al. 2018), (Tay, Luu, and Hui 2018), (Miao, Liu, and Gao 2019a), (Chen, Cui, et al. 2019), (Huang, Le Bras, et al. 2019), (Wang, Yu, Sun, et al. 2019), (Sharma and Roychowdhury 2019), (Sun et al. 2019), (Xia, Wu, and Yan 2019), (Yu, Zha, and Yin 2019), (Tu et al. 2019), (Tang, Cai, and Zhuo 2019), (Miao, Liu, and Gao 2019b), (Li, Zhang, Zhu, et al. 2019), (Guo et al. 2020), (Niu et al. 2020), (Zhou, Luo, and Wu 2020), (Zhang, Zhao, et al. 2020), (Zhang, Luo, et al. 2020), (Jin et al. 2020), (Song et al. 2020), (Zhang, Wu, et al. 2020) |
| | | CLOZE | (Yadav, Vig, and Shroff 2017), (Cui, Chen, et al. 2017), (Dhingra et al. 2017), (Hoang, Wiseman, and Rush 2018), (Dhingra, Pruthi, and Rajagopal 2018), (Yang, Dhingra, et al. 2017), (Seo et al. 2017), (Chen, Bolton, and Manning 2016), (Kobayashi et al. 2016), (Cui et al. 2016), (Kadlec, Bajgar, and Kleindienst 2016), (Kadlec et al. 2016), (Liu and Perez 2017), (Munkhdalai and Yu 2017), (Ma, Jurczyk, and Choi 2018), (Ghaeini et al. 2018), (Sugawara et al. 2018), (Mihaylov and Frank 2018), (Song et al. 2018), (Zhang, Zhao, et al. 2019), (Wang, Yao, et al. 2019), (Wang, Zhang, et al. 2020), (Niu et al. 2020), (Pappas et al. 2020) |
| | | DETAIL | (Chen et al. 2017), (Yang, Hu, et al. 2017), (Weissenborn, Wiese, and Seiffe 2017), (Hu, Peng, Huang, et al. 2018), (Shen et al. 2017), (Wang et al. 2017), (Liu, Shen, et al. 2018), (Huang et al. 2018), (Zhang et al. 2017), (Liu et al. 2017), (Gong and Bowman 2018), (Wang, Yuan, and Trischler 2017), (Golub et al. 2017), (Xiong, Zhong, and Socher 2018), (Xie and Xing 2017), (Salant and Berant 2018), (Mihaylov, Kozareva, and Frank 2017), (Kundu and Ng 2018b), (Prakash, Tripathy, and Banu 2018), (Kundu and Ng 2018a), (Hu, Peng, Wei, et al. 2018), (Seo et al. 2018), (Back et al. 2018), (Wang, Yan, and Wu 2018), (Clark and Gardner 2018), (Wang, Yu, Guo, et al. 2018), (Swayamdipta, Parikh, and Kwiatkowski 2018), (Wang, Yu, Jiang, et al. 2018), (Liu, Huang, et al. 2018), (Min, Seo, and Hajishirzi 2017), (Choi, Hewlett, Uszkoreit, et al. 2017), (Hewlett, Jones, and Lacoste 2017), (Bauer, Wang, and Bansal 2018), (Tan, Wei, Yang, et al. 2018), (Liu, Wei, et al. 2018),(Li, Li, and Lv 2018), (Wang, Liu, Liu, et al. 2018), (Aghaebrahimian 2018), (Dhingra, Pruthi, and Rajagopal 2018),(Yang, Dhingra, et al. 2017), (Seo et al. 2017), (Wang and Jiang 2017), (Trischler et al. 2017), (Xiong, Zhong, and Socher 2017), (Wang et al. 2016), (Liu and Perez 2017), , (Indurthi et al. 2018), (Min et al. 2018), (Sugawara et al. 2018), (Tay, Luu, and Hui 2018), (Gupta et al. 2018), (Ke et al. 2018), (Du and Cardie 2018), (Tan, Wei, Zhou, et al. 2018), (Wang, Liu, Xiao, et al. 2018), (Yu et al. 2018), (Nishida et al. 2018), (Wang and Bansal 2018), (Sachan and Xing 2018), (Htut, Bowman, and Cho 2018), (Yan et al. 2019), (Hu, Peng, et al. 2019a), (Wang, Gan, et al. 2019), (Jin, Yang, and Zhu 2019), (Angelidis et al. 2019), (Ding et al. 2019), (Takahashi et al. 2019), (Cui, Che, et al. 2019), (Mihaylov and Frank 2019), (Li, Zhang, Liu, et al. 2019), (Wang, Yu, Guo, et al. 2019), (Dua et al. 2019), (Zhang, Zhao, et al. 2019), (Xu, Liu, Chen, et al. 2019), (Yang, Wang, et al. 2019), (Wang and Jiang 2019), (Jiang et al. 2019), (Su et al. 2019), (Andor et al. 2019), (Pang et al. 2019), (Tang et al. 2019), (Wu et al. 2019), (Bi et al. 2019), (Dehghani et al. 2019), (Min et al. 2019), (Yuan et al. 2019), (Ren et al. 2019), (Nishida, Saito, et al. 2019), (Xu, Liu, Shen, et al. 2019), (Ran et al. 2019), (Li, Zhang, Zhu, et al. 2019), (Osama, El-Makky, and Torki 2019), (Liu, Zhang, Zhang, and Wang 2019), (Huang, Tang, et al. 2019), (Hu, Peng, et al. 2019b), (Nie, Wang, and Bansal 2019), (Tay et al. 2019), (Xiong et al. 2019), (Fremann 2019), (Hu, Wei, et al. 2019), (Zhuang and Wang 2019), (Li, Li, and Liu 2019), (Park, Lee, and Song 2019), (Nishida, Nishida, et al. 2019), (Lee, Kim, and Park 2019) , (Das et al. 2019), (Nakatsuji and Okui 2020), (Lee and Kim 2020), (Yang, Kang, and Seo 2020), (Wu and Xu 2020), (Zhang and Wang 2020), (Liu, Gong, et al. 2020), (Tu et al. 2020), (Ren, Cheng, and Su 2020), (Chen et al. 2020), (Niu et al. 2020), (Zheng et al. 2020), (Back et al. 2020), (Zhang, Wu, et al. 2020) |

**Table A2.** Reviewed papers categorized based on their embedding phase

| | | | | |
|---|---|---|---|---|
| Character embedding | CNN | | | (Clark and Gardner 2018), (Min, Seo, and Hajishirzi 2017), (Weissenborn, Wiese, and Seiffe 2017), (Shen et al. 2017), (Zhang et al. 2017), (Liu et al. 2017), (Gong and Bowman 2018), (Salant and Berant 2018), (Kundu and Ng 2018b), (Prakash, Tripathy, and Banu 2018), (Kundu and Ng 2018a), (Back et al. 2018), (Seo et al. 2018), (Seo et al. 2017), (Indurthi et al. 2018), (Sugawara et al. 2018), (Wang and Bansal 2018), (Yan et al. 2019), (Mihaylov and Frank 2019), (Zhang, Zhao, et al. 2019), (Xu, Liu, Chen, et al. 2019), (Wang and Jiang 2019), (Jiang et al. 2019), (Pang et al. 2019), (Tang et al. 2019), (Tang, Cai, and Zhuo 2019), (Yuan et al. 2019), (Ran et al. 2019), (Liu, Zhang, Zhang, and Wang 2019), (Huang, Tang, et al. 2019), (Xiong et al. 2019), (Zhuang and Wang 2019), (Park, Lee, and Song 2019), (Nishida, Nishida, et al. 2019), (Lee and Kim 2020), |
| | RNN | | | (Hu, Peng, Huang, et al. 2018),(Wang et al. 2017), (Tan, Wei, Yang, et al. 2018), (Wang, Yuan, and Trischler 2017), (Dhingra et al. 2017), (Prakash, Tripathy, and Banu 2018), (Hu, Peng, Wei, et al. 2018), (Dhingra, Pruthi, and Rajagopal 2018),(Yang, Dhingra, et al. 2017), (Ghaeini et al. 2018), (Gupta et al. 2018), (Du and Cardie 2018), (Tan, Wei, Zhou, et al. 2018),(Hu, Wei, et al. 2019), (Wu and Xu 2020) |
| | other | | | (Dua et al. 2019), (Xiong, Zhong, and Socher 2018), (Li, Li, and Lv 2018),(Wang, Yan, and Wu 2018),(Wang, Liu, Liu, et al. 2018),(Min et al. 2018), (Yu et al. 2018), (Yang, Kang, and Seo 2020), |
| Word embedding | Non-Contextual | One hot | | (Cui et al. 2016),(Liu and Perez 2017), (Tay, Luu, and Hui 2018), (Nishida et al. 2018), (Lee and Kim 2020), (Chen et al. 2020) |
| | | Learned | | (Choi, Hewlett, Uszkoreit, et al. 2017), (Greco et al. 2016), (Cui, Chen, et al. 2017), (Liu, Wei, et al. 2018), (Bauer, Wang, and Bansal 2018), (Yang, Dhingra, et al. 2017), (Cui et al. 2016), (Kadlec, Bajgar, and Kleindienst 2016), (Kadlec et al. 2016), (Ke et al. 2018), (Du and Cardie 2018), (Guo et al. 2020), (Chen et al. 2020), |
| | | Fixed Pre-train | | (Yan et al. 2019), (Angelidis et al. 2019), (Chen, Cui, et al. 2019), (Mihaylov and Frank 2019), (Dua et al. 2019), (Pang et al. 2019), (Wu et al. 2019), (Yu, Zha, and Yin 2019), (Tu et al. 2019), (Tang, Cai, and Zhuo 2019), (Yuan et al. 2019), (Xu, Liu, Shen, et al. 2019), (Ran et al. 2019), (Hu, Wei, et al. 2019), (Huang, Tang, et al. 2019), (Tay et al. 2019), (Xiong et al. 2019), (Zhuang and Wang 2019), (Wang, Yao, et al. 2019), (Park, Lee, and Song 2019), (Nishida, Nishida, et al. 2019), (Das et al. 2019), (Nakatsuji and Okui 2020), (Lee and Kim 2020), (Yang, Kang, and Seo 2020), (Wu and Xu 2020), (Song et al. 2020), (Back et al. 2020) |
| | | Fine-tune | | (Miao, Liu, and Gao 2019a), (Wang, Gan, et al. 2019), (Jin, Yang, and Zhu 2019) , (Zhang, Zhao, et al. 2019), (Xu, Liu, Chen, et al. 2019), (Wang and Jiang 2019), (Jiang et al. 2019), (Tang et al. 2019), (Bi et al. 2019), (Nishida, Saito, et al. 2019), (Miao, Liu, and Gao 2019b), (Liu, Zhang, Zhang, and Wang 2019) , (Wang, Zhang, et al. 2020), (Zhou, Luo, and Wu 2020), (Ren, Cheng, and Su 2020) |
| | Contextual | Learned | RNN | (Chen et al. 2017), (Min, Seo, and Hajishirzi 2017), (Weissenborn, Wiese, and Seiffe 2017), (Liu et al. 2017), (Hu, Peng, Huang, et al. 2018), (Liu, Shen, et al. 2018), (Huang et al. 2018), (Gong and Bowman 2018), (Wang, Yuan, and Trischler 2017), (Wang, Yu, Guo, et al. 2018), (Wang, Yu, Jiang, et al. 2018), (Golub et al. 2017), (Xiong, Zhong, and Socher 2018), (Xie and Xing 2017), (Salant and Berant 2018), (Mihaylov, Kozareva, and Frank 2017), (Wang, Yu, Chang, et al. 2018), (Kundu and Ng 2018b), (Liu, Wei, et al. 2018), (Kundu and Ng 2018a), (Hu, Peng, Wei, et al. 2018), (Bauer, Wang, and Bansal 2018), (Aghaebrahimian 2018), (Seo et al. 2018), (Li, Li, and Lv 2018), (Zhang et al. 2018), (Liu, Huang, et al. 2018), (Wang, Yan, and Wu 2018), (Wang, Liu, Liu, et al. 2018), (Liu, Zhao, et al. 2018), (Yang, Dhingra, et al. 2017), (Seo et al. 2017), (Wang and Jiang 2017), (Kobayashi et al. 2016), (Xiong, Zhong, and Socher 2017), (Wang et al. 2016), (Munkhdalai and Yu 2017), (Ma, Jurczyk, and Choi 2018), (Sugawara et al. 2018), (Ke et al. 2018), (Du and Cardie 2018), (Wang, Liu, Xiao, et al. 2018), (Nishida et al. 2018), (Wang and Bansal 2018), (Sachan and Xing 2018), (Yang, Hu, et al. 2017),(Clark and Gardner 2018), (Choi, Hewlett, Uszkoreit, et al. 2017),(Shen et al. 2017), (Wang et al. 2017), (Tan, Wei, Yang, et al. 2018), (Zhang et al. 2017), (Hewlett, Jones, and Lacoste 2017), (Greco et al. 2016), (Cui, Chen, et al. 2017), (Dhingra et al. 2017), (Prakash, Tripathy, and Banu 2018), (Lin, Liu, and Li 2018), (Hoang, Wiseman, and Rush 2018), (Back et al. 2018), (Dhingra, Pruthi, and Rajagopal 2018), (Chen, Bolton, and Manning 2016), (Trischler et al. 2017), (Cui et al. 2016), (Kadlec, Bajgar, and Kleindienst 2016), (Kadlec et al. 2016), (Indurthi et al. 2018), (Ghaeini et al. 2018), (Sheng, Lan, and Wu 2018), (Sugawara et al. 2018), (Tay, Luu, and Hui 2018), (Gupta et al. 2018), (Tan, Wei, Zhou, et al. 2018), (Mihaylov and Frank 2018), (Song et al. 2018), (Lee and Kim 2020), (Yang, Kang, and Seo 2020), (Nakatsuji and Okui |
| | | | | 2020), (Zhou, Luo, and Wu 2020), (Wu and Xu 2020) , (Ren, Cheng, and Su 2020), (Chen et al. 2020), (Song et al. 2020), (Zhang, Luo, et al. 2020), (Zhang and Wang 2020) |
| | | | CNN | (Yin, Ebert, and Schütze 2016), (Ma, Jurczyk, and Choi 2018), (Gupta et al. 2018), (Yu et al. 2018) |
| | | Fixed Pre-train | | (Wang, Yu, Sun, et al. 2019), (Xu, Liu, Shen, et al. 2019), (Hu, Wei, et al. 2019), (Lee and Kim 2020), (Wu and Xu 2020), (Back et al. 2020), (Zhang and Wang 2020), (Nakatsuji and Okui 2020) |
| | | Fine-tune | | (Hu, Peng, et al. 2019a), (Ding et al. 2019), (Takahashi et al. 2019), (Huang, Le Bras, et al. 2019), (Cui, Che, et al. 2019) , (Li, Zhang, Liu, et al. 2019), (Wang, Yu, Guo, et al. 2019), (Yang, Wang, et al. 2019), (Su et al. 2019), (Andor et al. 2019), (Sharma and Roychowdhury 2019), (Sun et al. 2019), (Xia, Wu, and Yan 2019), (Dehghani et al. 2019), (Min et al. 2019), (Ren et al. 2019), (Nishida, Saito, et al. 2019), (Li, Zhang, Zhu, et al. 2019), (Osama, El-Makky, and Torki 2019) , (Liu, Zhang, Zhang, and Wang 2019) , (Huang, Tang, et al. 2019), (Hu, Peng, et al. 2019b), (Nie, Wang, and Bansal 2019), (Frermann 2019), (Li, Li, and Liu 2019), (Lee, Kim, and Park 2019), (Liu, Gong, et al. 2020), (Back et al. 2020), (Zhang, Wu, et al. 2020), (Zhang, Zhao, et al. 2020), (Zhang, Luo, et al. 2020), (Tu et al. 2020), (Chen et al. 2020), (Zheng et al. 2020) |
| Hybrid | - | | | (Clark and Gardner 2018), (Min, Seo, and Hajishirzi 2017), (Weissenborn, Wiese, and Seiffe 2017), (Liu et al. 2017), (Hu, Peng, Huang, et al. 2018), (Shen et al. 2017), (Wang et al. 2017), (Tan, Wei, Yang, et al. 2018), (Zhang et al. 2017), (Gong and Bowman 2018), (Wang, Yuan, and Trischler 2017), (Xiong, Zhong, and Socher 2018), (Salant and Berant 2018), (Dhingra et al. 2017), (Kundu and Ng 2018b), (Prakash, Tripathy, and Banu 2018), (Kundu and Ng 2018a),(Hu, Peng, Wei, et al. 2018), (Back et al. 2018), (Seo et al. 2018), (Li, Li, and Lv 2018), (Wang, Yan, and Wu 2018), (Wang, Liu, Liu, et al. 2018), (Dhingra, Pruthi, and Rajagopal 2018), (Dhingra et al. 2017), (Yang, Dhingra, et al. 2017), (Seo et al. 2017), (Indurthi et al. 2018), (Ghaeini et al. 2018), (Min et al. 2018), (Sugawara et al. 2018), (Gupta et al. 2018), (Du and Cardie 2018), (Tan, Wei, Zhou, et al. 2018), (Yu et al. 2018), (Nishida et al. 2018), (Wang and Bansal 2018; Yan et al. 2019), (Mihaylov and Frank 2019), (Dua et al. 2019), (Zhang, Zhao, et al. 2019), (Xu, Liu, Chen, et al. 2019), (Tang, Cai, and Zhuo 2019), (Yuan et al. 2019), (Ran et al. 2019), (Liu, Zhang, Zhang, and Wang 2019), (Hu, Wei, et al. 2019), (Huang, Tang, et al. 2019), (Zhuang and Wang 2019), (Park, Lee, and Song 2019), (Nishida, Nishida, et al. 2019), (Wang and Jiang 2019), (Jiang et al. 2019), (Pang et al. 2019), (Tang et al. 2019), (Xiong et al. 2019), (Lee and Kim 2020), (Yang, Kang, and Seo 2020), (Wu and Xu 2020) |
| Sentence embedding | - | | | (Yin, Ebert, and Schütze 2016), (Liu and Perez 2017), (Chaturvedi, Pandit, and Garain 2018), (Min et al. 2018), (Htut, Bowman, and Cho 2018), (Wang, Yao, et al. 2019), (Park, Lee, and Song 2019), (Guo et al. 2020), (Zhou, Luo, and Wu 2020), (Tu et al. 2020), (Jin et al. 2020) |

**Table A3.** Reviewed papers categorized based on their reasoning phase

| | | |
|---|---|---|
| Direction | One-direction | (Chen et al. 2017), (Yang, Hu, et al. 2017), (Weissenborn, Wiese, and Seiffe 2017), (Shen et al. 2017), (Wang et al. 2017), (Huang et al. 2018), (Tan, Wei, Yang, et al. 2018), (Wang, Yu, Guo, et al. 2018), (Hewlett, Jones, and Lacoste 2017), (Xie and Xing 2017), (Salant and Berant 2018), (Wang, Yu, Jiang, et al. 2018), (Yadav, Vig, and Shroff 2017), (Dhingra et al. 2017), (Wang, Yu, Chang, et al. 2018), (Kundu and Ng 2018b), (Prakash, Tripathy, and Banu 2018), (Lin, Liu, and Li 2018), (Liu, Huang, et al. 2018), (Liu, Zhao, et al. 2018), (Wang and Jiang 2017), (Chen, Bolton, and Manning 2016), (Kobayashi et al. 2016), (Cui et al. 2016), (Yin, Ebert, and Schütze 2016), (Wang et al. 2016), (Kadlec, Bajgar, and Kleindienst 2016), (Kadlec et al. 2016), (Choi, Hewlett, Lacoste, et al. 2017), (Chaturvedi, Pandit, and Garain 2018), (Ghaeini et al. 2018), (Sheng, Lan, and Wu 2018), (Min et al. 2018), (Sugawara et al. 2018), (Gupta et al. 2018), (Ke et al. 2018), (Tan, Wei, Zhou, et al. 2018), (Wang, Liu, Xiao, et al. 2018), (Mihaylov and Frank 2018), (Song et al. 2018), (Sachan and Xing 2018), (Htut, Bowman, and Cho 2018), (Liu and Perez 2017), (Yan et al. 2019), (Wang, Gan, et al. 2019), (Angelidis et al. 2019), (Chen, Cui, et al. 2019), (Zhang, Zhao, et al. 2019), (Xu, Liu, Shen, et al. 2019), (Tay et al. 2019), (Wang, Zhang, et al. 2020), (Zhang and Wang 2020), (Ren, Cheng, and Su 2020), (Song et al. 2020) |
| | Two-direction | (Clark and Gardner 2018), (Min, Seo, and Hajishirzi 2017), (Liu et al. 2017), (Golub et al. 2017), (Swayamdipta, Parikh, and Kwiatkowski 2018), (Duan et al. 2017), (Cui, Chen, et al. 2017), (Liu, Wei, et al. 2018), (Zhang et al. 2018), (Aghaebrahimian 2018), (Back et al. 2018), (Seo et al. 2018), (Wang, Liu, Liu, et al. 2018), (Hu, Peng, Huang, et al. 2018), (Greco et al. 2016), (Zhang et al. 2017), (Gong and Bowman 2018), (Xiong, Zhong, and Socher 2018), (Kundu and Ng 2018a), (Hu, Peng, Wei, et al. 2018), (Bauer, Wang, and Bansal 2018), (Li, Li, and Lv 2018), (Wang, Yan, and Wu 2018), (Dhingra, Pruthi, and Rajagopal 2018), (Yang, Dhingra, et al. 2017), (Seo et al. 2017), (Xiong, Zhong, and Socher 2017), (Indurthi et al. 2018), (Sugawara et al. 2018), (Tay, Luu, and Hui 2018), (Yu et al. 2018), (Nishida et al. 2018), (Wang and Bansal 2018), (Miao, Liu, and Gao 2019a), (Hu, Peng, et al. 2019a), (Jin, Yang, and Zhu 2019), (Ding et al. 2019), (Takahashi et al. 2019), (Huang, Le Bras, et al. 2019), (Cui, Che, et al. 2019), (Mihaylov and Frank 2019), (Li, Zhang, Liu, et al. 2019), (Wang, Yu, Guo, et al. 2019), (Dua et al. 2019), (Zhang, Zhao, et al. 2019), (Xu, Liu, Chen, et al. 2019), (Yang, Wang, et al. 2019), (Wang, Yu, Sun, et al. 2019), (Wang and Jiang 2019), (Jiang et al. 2019), (Su et al. 2019), (Andor et al. 2019), (Pang et al. 2019), (Tang et al. 2019), (Sharma and Roychowdhury 2019), (Sun et al. 2019), (Wu et al. 2019), (Bi et al. 2019), (Xia, Wu, and Yan 2019), (Yu, Zha, and Yin 2019), (Dehghani et al. 2019), (Min et al. 2019), (Tu et al. 2019), (Tang, Cai, and Zhuo 2019), (Yuan et al. 2019), (Ren et al. 2019), (Das et al. 2019), (Nishida, Saito, et al. 2019), (Ran et al. 2019), (Miao, Liu, and Gao 2019b), (Li, Zhang, Zhu, et al. 2019), (Osama, El-Makky, and Torki 2019), (Liu, Zhang, Zhang, and Wang 2019), (Huang, Tang, et al. 2019), (Hu, Wei, et al. 2019), (Hu, Peng, et al. 2019b), (Nie, Wang, and Bansal 2019), (Xiong et al. 2019), (Frermann 2019), (Zhuang and Wang 2019), (Li, Li, and Liu 2019), (Park, Lee, and Song 2019), (Nishida, Nishida, et al. 2019), (Lee, Kim, and Park 2019), (Guo et al. 2020), (Niu et al. 2020), (Nakatsuji and Okui 2020), (Zhou, Luo, and Wu 2020), (Zhang, Zhao, et al. 2020), (Lee and Kim 2020), (Yang, Kang, and Seo 2020), (Wu and Xu 2020), (Liu, Gong, et al. 2020), (Tu et al. 2020), (Jin et al. 2020), (Pappas et al. 2020), (Zheng et al. 2020), (Chen et al. 2020), (Back et al. 2020), (Zhang, Wu, et al. 2020) , (Zhang, Luo, et al. 2020) |
| Dimension | One-dimension | (Weissenborn, Wiese, and Seiffe 2017), (Shen et al. 2017), (Tan, Wei, Yang, et al. 2018), (Hewlett, Jones, and Lacoste 2017), (Lin, Liu, and Li 2018), (Chen, Bolton, and Manning 2016), (Kobayashi et al. 2016), (Yin, Ebert, and Schütze 2016), (Wang et al. 2016), (Kadlec, Bajgar, and Kleindienst 2016), (Kadlec et al. 2016) , (Chaturvedi, Pandit, and Garain 2018), (Sheng, Lan, and Wu 2018), (Min et al. 2018), (Sugawara et al. 2018), (Ke et al. 2018), (Tan, Wei, Zhou, et al. 2018), (Mihaylov and Frank 2018), (Sachan and Xing 2018) , (Liu and Perez 2017), (Angelidis et al. 2019), (Tu et al. 2020), (Niu et al. 2020), (Ren, Cheng, and Su 2020), (Song et al. 2020) |
| | Two-dimension | (Chen et al. 2017), (Yang, Hu, et al. 2017), (Clark and Gardner 2018), (Min, Seo, and Hajishirzi 2017), (Liu et al. 2017), (Golub et al. 2017), (Swayamdipta, Parikh, and Kwiatkowski 2018), (Duan et al. 2017), (Cui, Chen, et al. 2017), (Liu, Wei, et al. 2018), (Zhang et al. 2018), (Aghaebrahimian 2018) ,(Back et al. 2018), (Seo et al. 2018), (Wang, Liu, Liu, et al. 2018), (Wang et al. 2017), (Wang, Yu, Guo, et al. 2018), (Salant and Berant 2018), (Wang, Yu, Jiang, et al. 2018), (Kundu and Ng 2018b), (Prakash, Tripathy, and Banu 2018), (Liu, Huang, et al. 2018), (Liu, Zhao, et al. 2018), (Hu, Peng, Huang, et al. 2018), (Lai et al. 2017), (Greco et al. 2016), (Dhingra et al. 2017), (Zhang et al. 2017), (Gong and Bowman 2018), (Kundu and Ng 2018a), (Hu, Peng, Wei, et al. 2018), (Bauer, Wang, and Bansal 2018), (Li, Li, and Lv 2018), (Wang, Yan, and Wu 2018), (Dhingra, Pruthi, and Rajagopal 2018), (Huang et al. 2018), (Wang, Yu, Chang, et al. 2018), (Xiong, Zhong, and Socher 2018), (Xie and Xing 2017), (Yadav, Vig, and Shroff 2017), (Yang, Dhingra, et al. 2017), (Seo et al. 2017), (Wang and Jiang 2017), (Xiong, Zhong, and Socher 2017), (Cui et al. 2016), (Indurthi et al. 2018), (Ghaeini et al. 2018), (Sugawara et al. 2018), (Tay, Luu, and Hui 2018), (Gupta et al. 2018), (Ke et al. 2018), (Wang, Liu, Xiao, et al. 2018), (Yu et al. 2018), (Song et al. 2018), (Nishida et al. 2018), (Wang and Bansal 2018), (Htut, Bowman, and Cho 2018), (Miao, Liu, and Gao 2019a), (Yan et al. |
| | | 2019), (Hu, Peng, et al. 2019a), (Wang, Gan, et al. 2019), (Jin, Yang, and Zhu 2019), (Ding et al. 2019), (Chen, Cui, et al. 2019), (Takahashi et al. 2019), (Huang, Le Bras, et al. 2019), (Cui, Che, et al. 2019), (Mihaylov and Frank 2019), (Li, Zhang, Liu, et al. 2019), (Wang, Yu, Guo, et al. 2019), (Dua et al. 2019), (Zhang, Zhao, et al. 2019) (Xu, Liu, Chen, et al. 2019), (Yang, Wang, et al. 2019), (Wang, Yu, Sun, et al. 2019), (Wang and Jiang 2019) (Jiang et al. 2019), (Su et al. 2019) , (Andor et al. 2019), (Pang et al. 2019), (Tang et al. 2019), (Sharma and Roychowdhury 2019), (Sun et al. 2019) , (Wu et al. 2019), (Bi et al. 2019), (Xia, Wu, and Yan 2019), (Yu, Zha, and Yin 2019), (Dehghani et al. 2019), (Min et al. 2019), (Tu et al. 2019), (Tang, Cai, and Zhuo 2019), (Yuan et al. 2019), (Ren et al. 2019), (Das et al. 2019), (Nishida, Saito, et al. 2019), (Xu, Liu, Shen, et al. 2019), (Ran et al. 2019), (Miao, Liu, and Gao 2019b), (Li, Zhang, Zhu, et al. 2019), (Osama, El-Makky, and Torki 2019), (Liu, Zhang, Zhang, and Wang 2019), (Hu, Wei, et al. 2019), (Huang, Tang, et al. 2019), (Hu, Peng, et al. 2019b) , (Nie, Wang, and Bansal 2019), (Tay et al. 2019), (Xiong et al. 2019) , (Fremann 2019), (Zhuang and Wang 2019), (Li, Li, and Liu 2019), (Park, Lee, and Song 2019), (Nishida, Nishida, et al. 2019), (Lee, Kim, and Park 2019), (Guo et al. 2020), (Wang, Zhang, et al. 2020), (Nakatsuji and Okui 2020), (Zhou, Luo, and Wu 2020), (Zhang, Zhao, et al. 2020), (Lee and Kim 2020), (Yang, Kang, and Seo 2020), (Wu and Xu 2020), (Zhang and Wang 2020), (Zhang, Luo, et al. 2020), (Liu, Gong, et al. 2020), (Jin et al. 2020), (Pappas et al. 2020), (Zheng et al. 2020), (Chen et al. 2020), (Back et al. 2020), (Zhang, Wu, et al. 2020) |
| NUMBER OF STEPS | SINGLE | (Chen et al. 2017), (Clark and Gardner 2018), (Min, Seo, and Hajishirzi 2017), (Huang et al. 2018), (Zhang et al. 2017), (Liu et al. 2017), (Golub et al. 2017), (Swayamdipta, Parikh, and Kwiatkowski 2018), (Duan et al. 2017), (Cui, Chen, et al. 2017), (Liu, Wei, et al. 2018), (Zhang et al. 2018), (Aghaebrahimian 2018), (Back et al. 2018), (Seo et al. 2018), (Wang, Liu, Liu, et al. 2018), (Weissenborn, Wiese, and Seiffe 2017), (Tan, Wei, Yang, et al. 2018), (Hewlett, Jones, and Lacoste 2017), (Wang et al. 2017), (Wang, Yu, Guo, et al. 2018), (Salant and Berant 2018), (Wang, Yu, Jiang, et al. 2018), (Kundu and Ng 2018b), (Prakash, Tripathy, and Banu 2018), (Liu, Huang, et al. 2018), (Liu, Zhao, et al. 2018), (Xiong, Zhong, and Socher 2018), (Xie and Xing 2017), (Yadav, Vig, and Shroff 2017), (Yang, Dhingra, et al. 2017), (Seo et al. 2017), (Wang and Jiang 2017), (Chen, Bolton, and Manning 2016), (Kobayashi et al. 2016), (Xiong, Zhong, and Socher 2017), (Cui et al. 2016), (Yin, Ebert, and Schütze 2016), (Wang et al. 2016), (Kadlec, Bajgar, and Kleindienst 2016), (Kadlec et al. 2016), (Choi, Hewlett, Lacoste, et al. 2017), (Chaturvedi, Pandit, and Garain 2018), (Indurthi et al. 2018), (Ghaeini et al. 2018), (Sheng, Lan, and Wu 2018), (Min et al. 2018), (Sugawara et al. 2018), (Tay, Luu, and Hui 2018), (Gupta et al. 2018), (Ke et al. 2018), (Tan, Wei, Zhou, et al. 2018), (Wang, Liu, Xiao, et al. 2018), (Mihaylov and Frank 2018), (Yu et al. 2018), (Nishida et al. 2018), (Wang and Bansal 2018), (Sachan and Xing 2018), (Htut, Bowman, and Cho 2018), (Miao, Liu, and Gao 2019a), (Wang, Gan, et al. 2019), (Jin, Yang, and Zhu 2019), (Chen, Cui, et al. 2019), (Zhang, Zhao, et al. 2019), (Xu, Liu, Chen, et al. 2019), (Wang and Jiang 2019), (Jiang et al. 2019), (Pang et al. 2019), (Tang et al. 2019), (Wu et al. 2019), (Bi et al. 2019), (Tu et al. 2019), (Yuan et al. 2019), (Xu, Liu, Shen, et al. 2019), (Ran et al. 2019), (Miao, Liu, and Gao 2019b), (Osama, El-Makky, and Torki 2019), (Liu, Zhang, Zhang, and Wang 2019), [8], (Tay et al. 2019), (Xiong et al. 2019), (Zhuang and Wang 2019), (Park, Lee, and Song 2019), (Nishida, Nishida, et al. 2019), (Niu et al. 2020), (Nakatsuji and Okui 2020), (Zhou, Luo, and Wu 2020), (Zhang, Zhao, et al. 2020), (Lee and Kim 2020), (Yang, Kang, and Seo 2020), (Wu and Xu 2020), (Zhang and Wang 2020), (Tu et al. 2020), (Ren, Cheng, and Su 2020), (Song et al. 2020), (Chen et al. 2020) |
| | MULTI- FIXED | (Yang, Hu, et al. 2017), (Lin, Liu, and Li 2018), (Hu, Peng, Huang, et al. 2018), (Greco et al. 2016), (Dhingra et al. 2017), (Gong and Bowman 2018), (Kundu and Ng 2018a), (Hu, Peng, Wei, et al. 2018), (Bauer, Wang, and Bansal 2018), (Li, Li, and Lv 2018), (Wang, Yan, and Wu 2018), (Dhingra, Pruthi, and Rajagopal 2018), (Wang, Yu, Chang, et al. 2018), (Liu and Perez 2017), (Yan et al. 2019), (Hu, Peng, et al. 2019a), (Angelidis et al. 2019), (Takahashi et al. 2019), (Huang, Le Bras, et al. 2019), (Cui, Che, et al. 2019), (Mihaylov and Frank 2019), (Li, Zhang, Liu, et al. 2019), (Wang, Yu, Guo, et al. 2019), (Dua et al. 2019), (Zhang, Zhao, et al. 2019), (Yang, Wang, et al. 2019), (Wang, Yu, Sun, et al. 2019), (Su et al. 2019), (Andor et al. 2019), (Sharma and Roychowdhury 2019), (Sun et al. 2019), (Xia, Wu, and Yan 2019), (Dehghani et al. 2019), (Min et al. 2019), (Tang, Cai, and Zhuo 2019), (Ren et al. 2019), (Das et al. 2019), (Nishida, Saito, et al. 2019), (Li, Zhang, Zhu, et al. 2019), (Hu, Wei, et al. 2019), (Huang, Tang, et al. 2019), (Hu, Peng, et al. 2019b), (Nie, Wang, and Bansal 2019), (Fremann 2019), (Li, Li, and Liu 2019), (Lee, Kim, and Park 2019), (Guo et al. 2020), (Wang, Zhang, et al. 2020), (Pappas et al. 2020), (Zheng et al. 2020), (Jin et al. 2020), (Back et al. 2020), (Liu, Gong, et al. 2020), (Zhang, Wu, et al. 2020), (Zhang, Luo, et al. 2020) |
| | MULTI-DYNAMIC | (Shen et al. 2017), (Song et al. 2018), (Ding et al. 2019), (Yu, Zha, and Yin 2019) |

**Table A4.** Reviewed papers categorized based on their prediction phase

| | | |
|---|---|---|
| Extraction mode | Boundary identification | (Chen et al. 2017), (Yang, Hu, et al. 2017), (Clark and Gardner 2018), (Weissenborn, Wiese, and Seiffe 2017), (Hu, Peng, Huang, et al. 2018), (Zhang et al. 2017), (Liu et al. 2017), (Gong and Bowman 2018), (Wang, Yuan, and Trischler 2017), (Wang, Yu, Guo, et al. 2018), (Golub et al. 2017), (Swayamdipta, Parikh, and Kwiatkowski 2018), (Xiong, Zhong, and Socher 2018), (Xie and Xing 2017), (Salant and Berant 2018), (Wang, Yu, Jiang, et al. 2018), (Mihaylov, Kozareva, and Frank 2017), (Cui, Chen, et al. 2017), (Dhingra et al. 2017), (Kundu and Ng 2018b), (Liu, Wei, et al. 2018), (Kundu and Ng 2018a), (Hu, Peng, Wei, et al. 2018), (Hoang, Wiseman, and Rush 2018), (Back et al. 2018), (Seo et al. 2018), (Khashabi et al. 2018), (Li, Li, and Lv 2018), (Liu, Huang, et al. 2018), (Wang, Yan, and Wu 2018), (Wang, Liu, Liu, et al. 2018), (Dhingra, Pruthi, and Rajagopal 2018), (Shen et al. 2017), (Wang et al. 2017), (Liu, Shen, et al. 2018), (Huang et al. 2018), (Yang, Dhingra, et al. 2017), (Seo et al. 2017), (Wang and Jiang 2017), (Trischler et al. 2017), (Xiong, Zhong, and Socher 2017), (Cui et al. 2016), (Kadlec, Bajgar, and Kleindienst 2016), (Kadlec et al. 2016), (Munkhdalai and Yu 2017), (Ghaeini et al. 2018), (Min et al. 2018), (Sugawara et al. 2018), (Tay, Luu, and Hui 2018), (Gupta et al. 2018), (Ke et al. 2018), (Tan, Wei, Zhou, et al. 2018), (Mihaylov and Frank 2018), (Yu et al. 2018), (Nishida et al. 2018), (Htut, Bowman, and Cho 2018), (Yan et al. 2019), (Hu, Peng, et al. 2019a), (Wang, Gan, et al. 2019), (Jin, Yang, and Zhu 2019), (Ding et al. 2019), (Takahashi et al. 2019), (Cui, Che, et al. 2019), (Mihaylov and Frank 2019), (Li, Zhang, Liu, et al. 2019), (Wang, Yu, Guo, et al. 2019), (Dua et al. 2019), (Zhang, Zhao, et al. 2019), (Xu, Liu, Chen, et al. 2019), (Yang, Wang, et al. 2019), (Wang and Jiang 2019), (Su et al. 2019) , (Andor et al. 2019), (Pang et al. 2019), (Tang et al. 2019), (Wu et al. 2019), (Min et al. 2019), (Yuan et al. 2019), (Ren et al. 2019), (Das et al. 2019), (Xu, Liu, Shen, et al. 2019), (Ran et al. 2019), (Li, Zhang, Zhu, et al. 2019), (Osama, El-Makky, and Torki 2019), (Liu, Zhang, Zhang, and Wang 2019), (Hu, Wei, et al. 2019), (Huang, Tang, et al. 2019), (Hu, Peng, et al. 2019b), (Nie, Wang, and Bansal 2019), (Xiong et al. 2019), (Frermann 2019) , (Zhuang and Wang 2019), (Li, Li, and Liu 2019), (Park, Lee, and Song 2019), (Nishida, Nishida, et al. 2019), (Lee, Kim, and Park 2019), (Lee and Kim 2020), (Yang, Kang, and Seo 2020), (Wu and Xu 2020), (Zhang and Wang 2020), (Liu, Gong, et al. 2020), (Tu et al. 2020), (Ren, Cheng, and Su 2020), (Niu et al. 2020), (Zheng et al. 2020), (Chen et al. 2020), (Back et al. 2020), (Zhang, Wu, et al. 2020) |
| | Candidate ranking | (Min, Seo, and Hajishirzi 2017), (Choi, Hewlett, Uszkoreit, et al. 2017), (Shen et al. 2017), (Golub et al. 2017), (Duan et al. 2017), (Yadav, Vig, and Shroff 2017), (Prakash, Tripathy, and Banu 2018), (Aghaebrahimian 2018), (Chen, Bolton, and Manning 2016), (Kobayashi et al. 2016), (Yin, Ebert, and Schütze 2016), (Wang et al. 2016), (Liu and Perez 2017) , (Ma, Jurczyk, and Choi 2018), (Chaturvedi, Pandit, and Garain 2018), (Sheng, Lan, and Wu 2018), (Tay, Luu, and Hui 2018), (Du and Cardie 2018), (Wang, Liu, Xiao, et al. 2018), (Song et al. 2018), (Sachan and Xing 2018), (Angelidis et al. 2019), (Jiang et al. 2019), (Tu et al. 2019), (Guo et al. 2020), (Wang, Zhang, et al. 2020), (Ren, Cheng, and Su 2020), (Pappas et al. 2020), (Song et al. 2020) |
| Generation mode | Answer generation | (Hewlett, Jones, and Lacoste 2017), (Bauer, Wang, and Bansal 2018), (Tan, Wei, Yang, et al. 2018), (Indurthi et al. 2018), (Saha et al. 2018), (Ke et al. 2018), (Hu, Peng, et al. 2019a), (Dua et al. 2019), (Andor et al. 2019), (Bi et al. 2019), (Dehghani et al. 2019), (Nishida, Saito, et al. 2019), (Ran et al. 2019), (Tay et al. 2019), (Wang, Yao, et al. 2019), (Nakatsuji and Okui 2020), (Zhou, Luo, and Wu 2020), (Liu, Gong, et al. 2020), (Tu et al. 2020), (Zheng et al. 2020), (Chen et al. 2020), (Zhang, Wu, et al. 2020) |
| | Candidate ranking | (Greco et al. 2016), (Wang, Yu, Chang, et al. 2018), (Lin, Liu, and Li 2018), (Zhang et al. 2018), (Liu, Zhao, et al. 2018), (Miao, Liu, and Gao 2019a), (Chen, Cui, et al. 2019), (Huang, Le Bras, et al. 2019), (Wang, Yu, Sun, et al. 2019), (Sharma and Roychowdhury 2019), (Sun et al. 2019), (Xia, Wu, and Yan 2019), (Yu, Zha, and Yin 2019), (Tang, Cai, and Zhuo 2019), (Miao, Liu, and Gao 2019b), (Li, Zhang, Zhu, et al. 2019), (Zhang, Zhao, et al. 2020), (Zhang, Luo, et al. 2020), (Jin et al. 2020), (Niu et al. 2020), (Zhang, Wu, et al. 2020) |

**Table A5.** Reviewed papers categorized based on their evaluation metric

| | | |
|---|---|---|
| Extractive Metric | EM | (Chen et al. 2017), (Yang, Hu, et al. 2017), (Clark and Gardner 2018), (Joshi et al. 2017), (Weissenborn, Wiese, and Seiffe 2017), (Hu, Peng, Huang, et al. 2018), (Shen et al. 2017), (Wang et al. 2017), (Liu, Shen, et al. 2018), (Huang et al. 2018), (Liu et al. 2017), (Gong and Bowman 2018), (Wang, Yuan, and Trischler 2017), (Wang, Yu, Guo, et al. 2018), (Golub et al. 2017), (Swayamdipta, Parikh, and Kwiatkowski 2018), (Xie and Xing 2017), (Salant and Berant 2018), (Wang, Yu, Jiang, et al. 2018), (Mihaylov, Kozareva, and Frank 2017), (Kundu and Ng 2018b), (Kundu and Ng 2018a), (Hu, Peng, Wei, et al. 2018), (Back et al. 2018), (Seo et al. 2018), (Wang, Yan, and Wu 2018), (Dhingra, Pruthi, and Rajagopal 2018), (Prakash, Tripathy, and Banu 2018), (Aghaebrahimian 2018), (Yang, Dhingra, et al. 2017), (Seo et al. 2017), (Wang and Jiang 2017), (Trischler et al. 2017), (Xiong, Zhong, and Socher 2017), (Min et al. 2018), (Gupta et al. 2018), (Ke et al. 2018), (Du and Cardie 2018), (Wang, Liu, Xiao, et al. 2018), (Yu et al. 2018), (Nishida et al. 2018), (Htut, Bowman, and Cho 2018), (Yan et al. 2019), (Hu, Peng, et al. 2019a), (Wang, Gan, et al. 2019), (Ding et al. 2019), (Takahashi et al. 2019), (Cui, Che, et al. 2019), (Li, Zhang, Liu, et al. 2019), (Wang, Yu, Guo, et al. 2019), (Dua et al. 2019), (Zhang, Zhao, et al. 2019), (Xu, Liu, Chen, et al. 2019), (Yang, Wang, et al. 2019), (Wang, Yu, Sun, et al. 2019), (Wang and Jiang 2019), (Su et al. 2019), (Andor et al. 2019), (Pang et al. 2019), (Tang et al. 2019), (Wu et al. 2019), (Yu, Zha, and Yin 2019), (Dehghani et al. 2019), (Yuan et al. 2019), (Das et al. 2019), (Xu, Liu, Shen, et al. 2019), (Ran et al. 2019), (Li, Zhang, Zhu, et al. 2019), (Osama, El-Makky, and Torki 2019), (Hu, Wei, et al. 2019), (Huang, Tang, et al. 2019), (Nie, Wang, and Bansal 2019), (Xiong et al. 2019), (Zhuang and Wang 2019), (Li, Li, and Liu 2019), (Park, Lee, and Song 2019), (Nishida, Nishida, et al. 2019), (Lee, Kim, and Park 2019), (Wang, Zhang, et al. 2020), (Niu et al. 2020), (Lee and Kim 2020), (Yang, Kang, and Seo 2020), (Wu and Xu 2020), (Zhang and Wang 2020), (Tu et al. 2020), (Ren, Cheng, and Su 2020), (Chen et al. 2020), (Zhang, Wu, et al. 2020) |
| | F1 | (Chen et al. 2017), (Yang, Hu, et al. 2017), (Clark and Gardner 2018), (Joshi et al. 2017), (Weissenborn, Wiese, and Seiffe 2017), (Hu, Peng, Huang, et al. 2018), (Shen et al. 2017), (Wang et al. 2017), (Liu, Shen, et al. 2018), (Huang et al. 2018), (Liu et al. 2017), (Gong and Bowman 2018), (Wang, Yuan, and Trischler 2017), (Wang, Yu, Guo, et al. 2018), (Hewlett, Jones, and Lacoste 2017), (Golub et al. 2017), (Swayamdipta, Parikh, and Kwiatkowski 2018), (Xie and Xing 2017), (Salant and Berant 2018), (Wang, Yu, Jiang, et al. 2018), (Mihaylov, Kozareva, and Frank 2017), (Kundu and Ng 2018b), (Kundu and Ng 2018a), (Hu, Peng, Wei, et al. 2018), (Back et al. 2018), (Seo et al. 2018), (Wang, Yan, and Wu 2018), (Dhingra, Pruthi, and Rajagopal 2018), (Prakash, Tripathy, and Banu 2018), (Aghaebrahimian 2018), (Yang, Dhingra, et al. 2017), (Seo et al. 2017), (Wang and Jiang 2017), (Trischler et al. 2017), (Xiong, Zhong, and Socher 2017), (Min et al. 2018), (Sugawara et al. 2018), (Tay, Luu, and Hui 2018), (Gupta et al. 2018), (Ke et al. 2018), (Du and Cardie 2018), (Tan, Wei, Zhou, et al. 2018), (Wang, Liu, Xiao, et al. 2018), (Yu et al. 2018), (Nishida et al. 2018), (Wang and Bansal 2018), (Htut, Bowman, and Cho 2018), (Yan et al. 2019), (Hu, Peng, et al. 2019a), (Wang, Gan, et al. 2019), (Ding et al. 2019), (Takahashi et al. 2019), (Cui, Che, et al. 2019), (Li, Zhang, Liu, et al. 2019), (Wang, Yu, Guo, et al. 2019), (Dua et al. 2019), (Zhang, Zhao, et al. 2019), (Xu, Liu, Chen, et al. 2019), (Yang, Wang, et al. 2019), (Wang, Yu, Sun, et al. 2019), (Wang and Jiang 2019), (Su et al. 2019), (Andor et al. 2019), (Pang et al. 2019), (Tang et al. 2019), (Wu et al. 2019), (Yu, Zha, and Yin 2019), (Dehghani et al. 2019), (Min et al. 2019), (Yuan et al. 2019), (Das et al. 2019), (Xu, Liu, Shen, et al. 2019), (Ran et al. 2019), (Li, Zhang, Zhu, et al. 2019), (Osama, El-Makky, and Torki 2019), (Hu, Wei, et al. 2019), (Huang, Tang, et al. 2019) , (Nie, Wang, and Bansal 2019), (Xiong et al. 2019), (Zhuang and Wang 2019), (Park, Lee, and Song 2019), (Nishida, Nishida, et al. 2019), (Lee, Kim, and Park 2019), (Wang, Zhang, et al. 2020), (Niu et al. 2020), (Lee and Kim 2020), (Yang, Kang, and Seo 2020), (Wu and Xu 2020), (Zhang and Wang 2020), (Liu, Gong, et al. 2020), (Tu et al. 2020), (Ren, Cheng, and Su 2020), (Chen et al. 2020), (Zheng et al. 2020), (Zhang, Wu, et al. 2020) |
| | MAP | (Min, Seo, and Hajishirzi 2017), (Duan et al. 2017), (Wang et al. 2016), (Min et al. 2018), (Sachan and Xing 2018) |
| | MRR | (Min, Seo, and Hajishirzi 2017), (Duan et al. 2017), (Wang et al. 2016), (Sachan and Xing 2018), (Angelidis et al. 2019) |
| | P@1 | (Min, Seo, and Hajishirzi 2017), (Du and Cardie 2018), (Tan, Wei, Zhou, et al. 2018), (Sachan and Xing 2018), (Angelidis et al. 2019), (Ding et al. 2019), (Wang, Zhang, et al. 2020), (Niu et al. 2020), (Liu, Gong, et al. 2020), (Zheng et al. 2020), (Tu et al. 2020) |
| | R@1 | (Liu, Huang, et al. 2018), (Du and Cardie 2018), (Tan, Wei, Zhou, et al. 2018), (Ding et al. 2019), (Wang, Zhang, et al. 2020), (Niu et al. 2020), (Liu, Gong, et al. 2020), (Zheng et al. 2020) |
| | ACC | (Choi, Hewlett, Uszkoreit, et al. 2017), (Lai et al. 2017), (Yadav, Vig, and Shroff 2017), (Cui, Chen, et al. 2017), (Dhingra et al. 2017), (Wang, Yu, Chang, et al. 2018), (Lin, Liu, and Li 2018), (Hoang, Wiseman, and Rush 2018), (Zhang et al. 2018), (Liu, Zhao, et al. 2018), (Prakash, Tripathy, and Banu 2018), (Duan et al. 2017), (Yang, Dhingra, et al. 2017), (Seo et al. 2017), (Chen, Bolton, and Manning 2016), (Kobayashi et al. 2016), (Cui et al. 2016), (Yin, Ebert, and Schütze 2016), (Kadlec, Bajgar, and Kleindienst 2016), (Kadlec et al. 2016), (Liu and Perez 2017), (Choi, Hewlett, Lacoste, et al. 2017), (Munkhdalai and Yu 2017), (Ma, Jurczyk, and Choi 2018), |
| | | (Chaturvedi, Pandit, and Garain 2018), (Ghaeini et al. 2018), (Sheng, Lan, and Wu 2018), (Min et al. 2018), (Sugawara et al. 2018), (Tay, Luu, and Hui 2018), (Tan, Wei, Zhou, et al. 2018), (Mihaylov and Frank 2018), (Song et al. 2018), (Miao, Liu, and Gao 2019a), (Chen, Cui, et al. 2019), (Huang, Le Bras, et al. 2019), (Zhang, Zhao, et al. 2019), (Wang, Yu, Sun, et al. 2019), (Jiang et al. 2019), (Andor et al. 2019), (Sharma and Roychowdhury 2019), (Sun et al. 2019), (Xia, Wu, and Yan 2019), (Yu, Zha, and Yin 2019), (Tu et al. 2019), (Tang, Cai, and Zhuo 2019), (Miao, Liu, and Gao 2019b), (Li, Zhang, Zhu, et al. 2019), (Wang, Yao, et al. 2019), (Guo et al. 2020), (Niu et al. 2020), (Pappas et al. 2020), (Zhang, Zhao, et al. 2020), (Zhang, Luo, et al. 2020), (Jin et al. 2020), (Pappas et al. 2020), (Song et al. 2020), (Tu et al. 2020), (Zhang, Wu, et al. 2020) |
| | Hit@k/Top@k | (Greco et al. 2016) |
| Generative Metric | ROUGE_L | (Weissenborn, Wiese, and Seiffe 2017), (Aghaebrahimian 2018), (Liu, Wei, et al. 2018), (Tan, Wei, Yang, et al. 2018), (Li, Li, and Lv 2018), (Wang, Liu, Liu, et al. 2018), (Bauer, Wang, and Bansal 2018), (Nguyen et al. 2016), (Indurthi et al. 2018), (Sugawara et al. 2018), (Tay, Luu, and Hui 2018), (Ke et al. 2018), (Yan et al. 2019), (Wang, Gan, et al. 2019), (Jin, Yang, and Zhu 2019), (Mihaylov and Frank 2019), (Bi et al. 2019), (Ren et al. 2019), (Nishida, Saito, et al. 2019), (Liu, Zhang, Zhang, and Wang 2019), (Tay et al. 2019), (Frermann 2019), (Li, Li, and Liu 2019), (Nakatsuji and Okui 2020), (Zhou, Luo, and Wu 2020) |
| | BLEU | (Weissenborn, Wiese, and Seiffe 2017), (Aghaebrahimian 2018), (Tan, Wei, Yang, et al. 2018), (Li, Li, and Lv 2018), (Wang, Liu, Liu, et al. 2018), (Bauer, Wang, and Bansal 2018), (Trischler et al. 2017), (Indurthi et al. 2018), (Tay, Luu, and Hui 2018), (Ke et al. 2018), (Yan et al. 2019), (Wang, Gan, et al. 2019), (Jin, Yang, and Zhu 2019), (Mihaylov and Frank 2019), (Bi et al. 2019), (Ren et al. 2019), (Nishida, Saito, et al. 2019), (Liu, Zhang, Zhang, and Wang 2019), (Tay et al. 2019), (Frermann 2019), (Wang, Zhang, et al. 2020), (Nakatsuji and Okui 2020), (Zhou, Luo, and Wu 2020) |
| | METEOR | (Bauer, Wang, and Bansal 2018), (Indurthi et al. 2018), (Tay, Luu, and Hui 2018), (Tay et al. 2019), (Frermann 2019) |
| | CIDEr | (Bauer, Wang, and Bansal 2018), (Trischler et al. 2017) |

**Table A6.** Reviewed papers categorized based on their novelties

| | | |
|---|---|---|
| MODEL STRUCTURE | INPUT/ OUTPUT | (Choi, Hewlett, Uszkoreit, et al. 2017), (Liu, Shen, et al. 2018), (Tan, Wei, Yang, et al. 2018), (Liu et al. 2017), (Hewlett, Jones, and Lacoste 2017), (Swayamdipta, Parikh, and Kwiatkowski 2018), (Salant and Berant 2018), (Wang, Yu, Jiang, et al. 2018), (Liu, Huang, et al. 2018), (Longpre et al. 2019), (Wang, Yu, Sun, et al. 2019), (Zhu et al. 2019), (Guo et al. 2020), (Lee and Kim 2020), (Zheng et al. 2020) |
| | INTERNAL | (Chen et al. 2017), (Clark and Gardner 2018), (Weissenborn, Wiese, and Seiffe 2017), (Hu, Peng, Huang, et al. 2018), (Shen et al. 2017), (Wang et al. 2017), (Lai et al. 2017), (Tan, Wei, Yang, et al. 2018), (Huang et al. 2018), (Gong and Bowman 2018), (Xiong, Zhong, and Socher 2018), (Xie and Xing 2017), (Yadav, Vig, and Shroff 2017), (Cui, Chen, et al. 2017), (Dhingra et al. 2017), (Wang, Yu, Chang, et al. 2018), (Kundu and Ng 2018b), (Prakash, Tripathy, and Banu 2018), (Lin, Liu, and Li 2018), (Liu, Wei, et al. 2018), (Kundu and Ng 2018a), (Bauer, Wang, and Bansal 2018), (Aghaebrahimian 2018), (Back et al. 2018), (Seo et al. 2018), (Wang, Yan, and Wu 2018), (Wang, Liu, Liu, et al. 2018), (Liu, Zhao, et al. 2018), (Zhang et al. 2018) ,(Giuseppe 2017), (Yang, Dhingra, et al. 2017), (Seo et al. 2017), (Wang and Jiang 2017), (Kobayashi et al. 2016), (Trischler et al. 2017), (Xiong, Zhong, and Socher 2017), (Cui et al. 2016), (Yin, Ebert, and Schütze 2016), (Wang et al. 2016), (Kadlec et al. 2016), (Liu and Perez 2017), (Munkhdalai and Yu 2017), (Ma, Jurczyk, and Choi 2018), (Chaturvedi, Pandit, and Garain 2018), (Indurthi et al. 2018), (Ghaeini et al. 2018), (Sheng, Lan, and Wu 2018), (Min et al. 2018), (Tay, Luu, and Hui 2018), (Gupta et al. 2018), (Ke et al. 2018), (Tan, Wei, Zhou, et al. 2018), (Wang, Liu, Xiao, et al. 2018), (Mihaylov and Frank 2018), (Yu et al. 2018), (Song et al. 2018), (Htut, Bowman, and Cho 2018), (Miao, Liu, and Gao 2019a), (Yan et al. 2019), (Hu, Peng, et al. 2019a), (Wang, Gan, et al. 2019), (Jin, Yang, and Zhu 2019), (Angelidis et al. 2019), (Ding et al. 2019), (Chen, Cui, et al. 2019), (Takahashi et al. 2019), (Huang, Le Bras, et al. 2019), (Cui, Che, et al. 2019), (Mihaylov and Frank 2019), (Wang, Yu, Guo, et al. 2019), (Dua et al. 2019), (Zhang, Zhao, et al. 2019), (Xu, Liu, Chen, et al. 2019), (Yang, Wang, et al. 2019), (Jiang et al. 2019), (Su et al. 2019), (Andor et al. 2019), (Pang et al. 2019), (Tang et al. 2019), (Sharma and Roychowdhury 2019), (Sun et al. 2019), (Wu et al. 2019), (Bi et al. 2019), (Xia, Wu, and Yan 2019), (Yu, Zha, and Yin 2019), (Dehghani et al. 2019), (Min et al. 2019), (Tu et al. 2019), (Tang, Cai, and Zhuo 2019), (Yuan et al. 2019), (Ren et al. 2019), (Das et al. 2019), (Nishida, Saito, et al. 2019), (Xu, Liu, Shen, et al. 2019), (Ran et al. 2019), (Miao, Liu, and Gao 2019b), (Li, Zhang, Zhu, et al. 2019), (Osama, El-Makky, and Torki 2019), (Liu, Zhang, Zhang, and Wang 2019), (Hu, Wei, et al. 2019), (Huang, Tang, et al. 2019), (Hu, Peng, et al. 2019b), (Nie, Wang, and Bansal 2019), (Tay et al. 2019), (Xiong et al. 2019), (Frermann 2019), (Zhuang and Wang 2019), (Wang, Yao, et al. 2019), (Park, Lee, and Song 2019), (Nishida, Nishida, et al. 2019), (Lee, Kim, and Park 2019), (Wang, Zhang, et al. 2020), (Yang, Kang, and Seo 2020), (Niu et al. 2020), (Nakatsuji and Okui 2020), (Zhou, Luo, and Wu 2020), (Zhang, Zhao, et al. 2020), (Wu and Xu 2020), (Zhang and Wang 2020), (Zhang, Luo, et al. 2020), (Liu, Gong, et al. 2020), (Tu et al. 2020), (Welbl et al. 2020), (Jin et al. 2020), (Ren, Cheng, and Su 2020), (Chen et al. 2020), (Pappas et al. 2020), (Zheng et al. 2020), (Song et al. 2020), (Asai and Hajishirzi 2020), (Yan et al. 2020), (Back et al. 2020), (Zhang, Wu, et al. 2020), (Cao et al. 2020) (Zhou, Huang, and Zhu 2020), |
| DATASET | | (Joshi et al. 2017), (Lai et al. 2017), (He et al. 2018), (Welbl, Liu, and Gardner 2017), (Choi et al. 2018), (Xie et al. 2018), (Pampari et al. 2018), (Khashabi et al. 2018), (Ostermann et al. 2018), (Nguyen et al. 2016), (Trischler et al. 2017), (Bajgar, Kadlec, and Kleindienst 2017), (Cui et al. 2016), (Rajpurkar et al. 2016), (Onishi et al. 2016), (Ma, Jurczyk, and Choi 2018), (Šuster and Daelemans 2018), (Shao et al. 2018), (Saha et al. 2018), (Elgohary, Zhao, and Boyd-Graber 2018), (Tan, Wei, Zhou, et al. 2018), (Rajpurkar, Jia, and Liang 2018), (Hill et al. 2016), (Kočiský et al. 2018), (Kwiatkowski et al. 2019), (Welbl, Stenetorp, and Riedel 2018), (Yang et al. 2018), (Liang, Li, and Yin 2019), (Cui, Liu, et al. 2019), (Gupta et al. 2019), (Hardalov, Koychev, and Nakov 2019), (Jing, Xiong, and Zhen 2019), (Huang, Le Bras, et al. 2019), (Dua et al. 2019), (Sayama, Araujo, and Fernandes 2019), (Ostermann, Roth, and Pinkal 2019), (Mozannar et al. 2019), (Dasigi et al. 2019), (Yao et al. 2019), (Li, Li, and Liu 2019), (Liu, Lin, et al. 2019), (Anuranjana, Rao, and Mamidi 2019), (Lin et al. 2019), (Pappas et al. 2020), (Watarai and Tsuchiya 2020), (Sun et al. 2020), (Inoue, Stenetorp, and Inui 2020), (Yu et al. 2020), (Wang, Yao, et al. 2020), (Lee, Hwang, and Cho 2020), (Berzak, Malmaud, and Levy 2020), (Yuan, Fu, et al. 2020), (Horbach et al. 2020) |
| KNOWLEDGE TRANSFER | | (Chen et al. 2017), (Yang, Hu, et al. 2017), (Min, Seo, and Hajishirzi 2017), (Wang, Yuan, and Trischler 2017), (Golub et al. 2017), (Duan et al. 2017), (Yadav, Vig, and Shroff 2017), (Mihaylov, Kozareva, and Frank 2017), (Hu, Peng, Wei, et al. 2018), (Hoang, Wiseman, and Rush 2018), (Wang, Liu, Liu, et al. 2018), (Dhingra, Pruthi, and Rajagopal 2018), (Liu, Huang, et al. 2018), (Yin, Ebert, and Schütze 2016), (Wang et al. 2016), (Kadlec, Bajgar, and Kleindienst 2016), (Du and Cardie 2018), (Nishida et al. 2018), (Sachan and Xing 2018), (Longpre et al. 2019), (Hardalov, Koychev, and Nakov 2019), (Cui, Che, et al. 2019), (Mihaylov and Frank 2019), (Li, Zhang, Liu, et al. 2019), (Yang, Wang, et al. 2019), (Wang and Jiang 2019), (Su et al. 2019), (Sun et al. 2019), (Wu et al. 2019), (Bi et al. 2019), (Xia, Wu, and Yan 2019), (Yu, Zha, and Yin 2019), (Qiu et al. 2019), (Xu, Liu, Shen, et al. 2019), (Nishida, Nishida, et al. 2019), (Wang, Lu, and Tang 2019), (Lee, Kim, and Park 2019), ("ORB: An Open Reading Benchmark for Comprehensive Evaluation of Machine Reading Comprehension" 2019), (Soni and Roberts 2020), (Wu and Xu 2020), (Tu et al. 2020), (Nishida et al. 2020), (Cao et al. 2020), (Zhou, Huang, and Zhu 2020), (Asai and Hajishirzi 2020), (Jin et al. 2020), (Yan et al. 2020), (Ren, Cheng, and Su 2020), (Dua, Singh, and Gardner 2020), (Yuan, Shou, et al. 2020), (Zhang, Wu, et al. 2020), (Gupta and Khade 2020) |

| | |
|---|---|
| EVALUATION MEASURE | (Jia and Liang 2017), (Sugawara et al. 2017), (Chen, Bolton, and Manning 2016), (Sugawara et al. 2018), (Tan, Wei, Zhou, et al. 2018), (Wang and Bansal 2018), (Pugaliya et al. 2019), (Wang, Yu, Guo, et al. 2019), (Chen, Stanovsky, et al. 2019), (Bao et al. 2019), (Lee et al. 2019), (Fisch et al. 2019), (Talmor and Berant 2019), (Gardner et al. 2019), (Li, Li, and Liu 2019), (Wu and Xu 2020), (Schlegel et al. 2020), (Liu, Liu, et al. 2020), (Sugawara et al. 2020), (Yue, Gutierrez, and Sun 2020), (Charlet et al. 2020), (Soni and Roberts 2020), (Inoue, Stenetorp, and Inui 2020), (Berzak, Malmaud, and Levy 2020), (Welbl et al. 2020), (Zhou, Huang, and Zhu 2020), (Ren, Cheng, and Su 2020), (Cao et al. 2020), (Dunietz et al. 2020), (Horbach et al. 2020) |